\documentclass{article} % For LaTeX2e
\usepackage{iclr2027_conference,times}

\usepackage{amsmath,amsfonts,bm}

\def\eqref#1{equation~\ref{#1}}
\def\1{\bm{1}}

\def\vb{{\bm{b}}}

\def\vh{{\bm{h}}}

\def\vv{{\bm{v}}}

\def\vz{{\bm{z}}}

\def\mW{{\bm{W}}}

\DeclareMathAlphabet{\mathsfit}{\encodingdefault}{\sfdefault}{m}{sl}
\SetMathAlphabet{\mathsfit}{bold}{\encodingdefault}{\sfdefault}{bx}{n}

\newcommand{\R}{\mathbb{R}}

\usepackage{hyperref}
\usepackage{url}
\usepackage{graphicx}
\usepackage{booktabs}
\usepackage{multirow}
\usepackage{xcolor}
\usepackage{threeparttable}
\graphicspath{{images/}}

\title{Clinical Concept Centers in LLMs}

\author{%
  \textbf{Aishik Nagar}\textsuperscript{4} \quad
  \textbf{Abhishek Vaidyanathan}\textsuperscript{1,5} \quad
  \textbf{Arun-Kumar Kaliya-Perumal}\textsuperscript{2} \\
  \textbf{Elijah Tzen Hsuen Boey}\textsuperscript{\textbf{2}} \quad
  \textbf{Stefan Winkler}\textsuperscript{\textbf{1,3}} \\[6pt]
  \textsuperscript{1}\ National University of Singapore \quad
  \textsuperscript{2}\ Nanyang Technological University \\
  \textsuperscript{3}\ Singapore Institute of Technology \quad
  \textsuperscript{4}\ Carnegie Mellon University (CMU) \\
  \textsuperscript{5}\ SAP Asia Pte Ltd
}

\iclrfinalcopy % Uncomment for camera-ready version, but NOT for submission.
\begin{document}

\maketitle
\lhead{Preprint. Work in Progress.}

\begin{abstract}
Large language models are increasingly used in clinical settings. However, research into the reliability and performance of these models has focused almost entirely on the language substrate, scoring what the model says, while the latent computations that produce these answers stay unexamined. At the same time, mechanistic interpretability has found that the latent space carries a higher fidelity of representation than the text: internal representations not only encode substantially more than the output verbalizes, but the stated reasoning also systematically omits features that causally drive the answer. An evaluation of model behavior in terms of mechanistic interpretability has not been explored in clinical decision support. We believe that it is a necessity, not only for improving model performance, but also to evaluate the safety and alignment of these models as medical devices. In this work, we extend behavioral evaluation into the latent space and ask whether clinical concepts exist as locatable, causally used representations inside open-weight LLMs. Reliable mechanistic interpretability in such settings uses dictionary learning over a labeled text corpus. We reformulate the EHR setting into exactly this approach, with clinical notes as the corpus and expert-assigned ICD billing codes as clinically validated proxies for automated learning of clinical concept centers.  We find dedicated clinical concept centers in the latent space of all eleven open models we test. These concept centers are interpretable, firing only on their aligned clinical narratives, and meaningfully and causally drive model behavior in both constrained and open-ended settings. These concept centers are not just analytical representations, but circuits that can be utilized in clinical practice, and we explore their use from the perspective of both evaluation and performance. From the evaluation standpoint, models stay internally coherent and keep using the relevant concept centers even under adversarial role-based priming, while aligned priming improves downstream clinical performance. From a performance perspective, we simulate realistic deployment settings and find that steering models along these centers leads to meaningful downstream improvements. 
Finally, we conduct a blinded clinician validation and find the activation and usage of these concept centers predicts clinicians preferences. 
%We find that among the cases where the concepts strongly separate two responses, prediction reaches 76\% agreement with clinician preference, regardless of how the model was prompted.

% Finally, we conduct a blinded clinician validation and find that where the concept centers are cleanly and strongly activated, their activation predicts the clinicians' preferences, reaching 76\% agreement on the quarter of cases where the centers separate the responses most strongly, regardless of how the model was prompted.
\end{abstract}

\section{Introduction}
\label{sec:intro}

Large language models are increasingly being used in clinical workflows, across settings from medical code prediction \citep{mullenbach2018explainable} and question answering over clinical documents \citep{kweon2024ehrnoteqa} to open-ended health conversations \citep{arora2025healthbench} and inpatient clinical decision support \citep{nagar2026clrvoyance}, with open medical foundation models released explicitly for clinical application development \citep{sellergren2025medgemma} and instruction-tuned models whose answers physicians have rated above physician-written ones on most clinical-utility axes \citep{singhal2023large,singhal2025toward}. The use of these models as a medical device demands a guarantee that when the model works through a medical problem, it does so in a way that is aligned with how clinicians reason about it, with no implicit computations in the latent space biasing its downstream suggestions. Behavior-oriented benchmarks alone cannot tell us when a model can be trusted. Despite performing well on evaluations in the language substrate, deployed LLMs produce no significant gain in physicians' diagnostic reasoning in a randomized trial \citep{goh2024large}, and models which perform well on exams fall short of clinicians on realistic EHR decisions \citep{hager2024evaluation}.

However, nearly all evaluation of LLMs in clinical settings happens on the language substrate, scoring what the model says as exam answers \citep{jin2021disease,pal2022medmcqa} or physician-rubric grades over generated conversations \citep{arora2025healthbench}. This is not a faithful window into the model's latent computation. Internal representations encode substantially more than the output verbalizes, to the point that a model can internally encode a correct answer while generating a wrong one \citep{orgad2025llms}, and stated chain-of-thought reasoning systematically omits the features that causally drive the answer \citep{turpin2023language,lanham2023measuring}. Because these biases or shortcuts may remain hidden in the LLM’s latent representations, output-level evaluations alone are insufficient to detect or prevent them, particularly in clinical settings. %, because an output-level evaluation cannot see a bias or a shortcut that lives only in the latent space. 
Interpretability of clinical models has itself largely stayed on the behavioral side, analyzing the language substrate, and mechanistic work in the latents is only beginning (\S\ref{sec:related}). No such work exists that identifies expert-label-grounded clinical concepts inside the model and tests whether they causally influence the model's clinical behavior.

Finding such representations, which we call \emph{clinical concept centers}, has a twofold advantage. From the evaluation perspective, black-box access is insufficient for rigorous audits because root-cause diagnosis needs the internals \citep{casper2024black}, regulators are moving oversight toward transparency about how a deployed model decides \citep{mesko2023imperative}, and our position is that for a model used as a medical device, evaluation in the mechanistic space is as necessary as evaluation in the language substrate. From the performance perspective, reliable concept representations can serve as steerable vectors. Prior work has already shown that interventions on internal representations edit and steer behavior without retraining \citep{meng2022locating,zou2023representation,rimsky2024steering}, reinforcement-learning post-training mostly amplifies reasoning paths already present in the base model \citep{yue2025does}, and there is an increasing body of work that uses mechanistic methods to find the reasoning circuits that develop during such post-training and introduce them into models at a fraction of the training cost, though so far by brute-force feature selection \citep{wang2025resa}. A well-informed mechanistic approach for improving model capabilities in clinical settings requires the right concept centers to be identified and confirmed to causally influence model behavior, which further motivates this study.

So the question becomes whether clinical concept centers even exist in these models, and how to locate them. Whatever the instrument for interpretable approaches over model latents, from dictionary learning \citep{templeton2026scaling,cunningham2024sparse} to latent probes \citep{alain2016understanding} to direction extraction on the raw residual stream \citep{marks2024geometry}, they are most reliable when they run over a text corpus with annotated labels, so that the recovered representations can be named by what they fire on. Automated interpretability can generate such labels with an LLM in the loop, but the generated explanations carry an error signal of their own \citep{singh2023explaining,paulo2024automatically}, which we want to avoid. Healthcare data already maintains this kind of labeled composition. The International Classification of Diseases (ICD) is the standard vocabulary in which diagnoses are recorded, and after essentially every patient encounter a trained medical coder reads the clinical documentation and assigns the codes describing what the encounter was about. The codes form a hierarchy, from thousands of highly specific leaf codes up to broad chapters corresponding to organ systems (circulatory, respiratory, musculoskeletal, and so on), and although their day-to-day purpose is record-keeping and billing, each assignment is an expert's judgment of the diagnostic concepts a note encodes, made at scale as routine clinical practice. Clinical ontologies like the ICD hierarchy are therefore established, expert-maintained terminologies, often used for representation learning in clinical foundation models \citep{choi2017gram}. EHR data such as discharge notes already carry these associated codes \citep{johnson2023mimic,PhysioNet-mimic-iv-note-2.2}, and the automated-coding literature uses exactly this setting as its ground truth \citep{mullenbach2018explainable}. We reformulate this EHR setting into the labeled-corpus recipe, with clinical notes as the text corpus and ICD codes as clinically validated labels, so instead of training a model on the ontology, as the foundation-model line does, we take an existing artifact of knowledge and ask whether its latent space already contains these representations. This formulation provides expert-annotated supervision at scale with no labeling effort, and has not been used for model interpretability before, which makes the formulation itself one of the contributions of this work.

% Maybe remove this paragraph -> we say the exact thing in the instrumentation method
%With the data formulation in place, the next choice is the instrumentation that probes the latent representations over it. Linear probes recover whatever is linearly decodable at a depth \citep{alain2016understanding}, but probe accuracy conflates what is encoded with what is used \citep{belinkov2022probing,hewitt2019designing,elazar2021amnesic}. Difference-of-means directions are the strongest simple carriers of a concept \citep{marks2024geometry}. Dictionary learning with sparse autoencoders (SAEs) recovers monosemantic features from polysemantic activations \citep{cunningham2024sparse,templeton2024scaling,gao2025scaling}, undoing the superposition that lets a model hold more concepts than dimensions \citep{elhage2022toy}, though a skeptical line argues SAEs underperform the simpler baselines as probes and as steering tools \citep{kantamneni2025are,wu2025axbench}. We investigate all three, probes, difference-of-means directions, and SAE-derived directions, and find that SAEs not only give the more robust mechanism for discovering a named, reusable set of features that can be monitored per generation step, but that the SAE-derived directions also empirically turn out to be the strongest lever for causally influencing the models (\S\ref{sec:steering}).

Thus, as the foundation, we establish existence, specificity, causal use, and temporal persistence of these concept centers across eleven open-weight Qwen and Gemma models (4B--30B, dense and mixture-of-experts, base, instruction-tuned, and medical variants) (\S\ref{sec:foundation}). 
We investigate the use of these concept centers in latent-space evaluation, performance improvements in clinical tasks and for predicting expert preference during clinician validation (\S\ref{sec:clinician}).
This leads to three key contributions:
% We probably need to explicitly say that we are the first to explore clinical interpretability in the latent space of models
% we could say something like its a less explored field and we are cnntributing blah blah?
\begin{itemize}
    \item \textbf{A novel, clinically grounded formulation for latent-space interpretability.} Reformulating EHR notes with expert-assigned ICD codes into a labeled corpus for dictionary learning keeps the probing setting realistic, provides expert-annotated supervision at scale without any labeling effort, and avoids the error signal that automated labeling introduces. The same recipe extends to any domain that maintains an established ontology over its records.
    \item \textbf{Dedicated, separable clinical concept centers in the latent space of open models.} We find the concept centers recoverable in all 11 models tested (best held-out single-feature AUC 0.80--0.92 in the Qwen family), responding to the clinical situation of a note rather than to its vocabulary, and transferring to other clinical settings with 45--86\% of excess-over-chance separability retained. These concept centers are used across the depth of the models and persistently through the generation of text for clinical support (\S\ref{sec:foundation}).
    \item \textbf{A meaningful demonstration of their use for evaluation and performance, validated with clinicians.} Counterfactual role-based priming becomes a latent-space evaluation of whether a model follows its assigned role or its own clinical concepts, showing that the models remain internally coherent under adversarial roles, while aligned priming improves accuracy by $+2.9$ to $+11.2$ points across both model families (\S\ref{sec:roles}). Steering the models along the concept centers under a realistic prior improves downstream clinical performance ($+1.1$ to $+2.3$ additional correct diagnoses per hundred cases on the Qwen models). A blinded clinician evaluation shows that the strength of the concept-center activations predict clinician preference, with a 76\% agreement on the largest activation gaps, regardless of how the model was prompted (\S\ref{sec:steering}--\S\ref{sec:clinician}).
\end{itemize}

\section{Related work}
\label{sec:related}

\paragraph{Clinical LLMs and clinical interpretability.}
Medical foundation models and instruction-tuned generalists reach expert-level scores on knowledge benchmarks \citep{singhal2023large,singhal2025toward,sellergren2025medgemma}. Evaluation has followed the outputs, from exam-style question answering \citep{jin2021disease,pal2022medmcqa,kweon2024ehrnoteqa} to physician-rubric grading of conversations \citep{arora2025healthbench}, and benchmark performance still fails to translate into clinician benefit or reliable decisions \citep{goh2024large,hager2024evaluation}. Interpretability has entered this domain only at the edges, decomposing radiology vision encoders \citep{abdulaal2024interpretable}, adapting dictionaries to clinical text \citep{o2025resurrecting}, examining reasoning variability \citep{modi2026clinical}, and localizing triage failures to output formatting \citep{navarro2026internal}. Understanding clinical model behavior has otherwise stayed on the behavioral side, analyzing the language substrate, and no existing latent-space work grounds the recovered features in expert-assigned diagnostic labels and tests them causally against the model's clinical behavior.

\paragraph{Dictionary learning, concept directions, and their critics.}
Concepts are often carried by linear directions \citep{park2024linear}, difference-of-means directions are causally implicated carriers \citep{marks2024geometry}, activation-level interventions steer behavior without retraining \citep{turner2024steering,rimsky2024steering,zou2023representation,meng2022locating}, and editing effectiveness dissociates from where probing localizes a concept \citep{hase2023does}, which our  results confirm in the clinical domain. A skeptical line reports that SAE probes underperform logistic regression \citep{kantamneni2025are}, that prompting and difference-of-means baselines beat SAE steering and detection \citep{wu2025axbench}, and that SAE-internal metrics do not guarantee downstream utility \citep{karvonen2025saebench}. We treat these as design constraints, keep the probe and the raw direction as baselines everywhere, follow the control-task logic of \citet{hewitt2019designing} and the erasure machinery of \citet{belrose2023leace} on the removal side, and report the one place our data contradicts the skepticism, the SAE-decoder direction being the stronger causal lever on this task (\S\ref{sec:steering}). Appendix~\ref{app:related} extends this section with the dictionary-caveat and steering-reliability lines that inform our controls.

\section{Method}
\label{sec:foundation}

% NOTE (space option): if the main text needs to shrink, causal-use-at-decision
% (\S 3.4) and persistence-through-generation (\S 3.5) can merge into one
% subsection, since persistence during generation subsumes use at the decision
% point; the extra decision-point ablations then move to the appendix.

% Add a different section for basic data and method since this is used across all other sections
\paragraph{Data and labels.}
We use MIMIC-IV discharge summaries \citep{johnson2023mimic,PhysioNet-mimic-iv-note-2.2}. Each summary contains ICD-9 diagnosis codes assigned by an expert medical coder. Two properties of these codes need consideration. Their ontology is extremely granular, the representations of model latents may not plausibly resolve concepts at that level of granularity, and the codes assigned to an encounter often include entries copied over from past occurrences or added for billing purposes, which adds its own source of error to the labels. 
We therefore use only the primary code of each note and group codes at the level of the 17 ICD-9-CM disease chapters, which are broad organ-system categories that behave like department specialties (mapping and counts in Appendix~\ref{app:prompt}). After patient-level deduplication and capping at 1{,}000 notes per chapter, the corpus holds 8{,}617 notes, split by patient into a selection half used for all direction estimation and hyperparameter choices and an evaluation half on which every reported number is computed (models of 27B parameters and above use a 4{,}334-note subset with the same balance properties). 
We study 11 open-weight models across two families, five Qwen models spanning base, chat, medical fine-tuned, and mixture-of-experts variants (8B--30B), and six Gemma models spanning pretrained, instruction-tuned, and medical variants (4B--27B), with dictionaries from the open Qwen and Gemma SAE suites \citep{deng2026qwen}. 

\paragraph{Constructing a concept center.}
Let $\vh_\ell(x) \in \R^{d}$ be the residual-stream activation of note $x$ at layer $\ell$, mean-pooled over the note's tokens (the final token of a decoder-only model encodes the next-token prediction rather than a summary of the input, so pooling is the faithful read of what the note activated). An SAE encodes $\vh$ into sparse features $\vz = f_{\text{enc}}(\vh) \in \R^{m}$, $m \gg d$, and decodes $\hat{\vh} = \mW_{\text{dec}}\vz + \vb_{\text{dec}}$ (TopK activation for the Qwen suite, JumpReLU \citep{rajamanoharan2024jumping} for the Gemma suite). For chapter $C$ we form the difference of mean sparse codes between selection notes inside and outside the chapter and decode it into model coordinates,
\begin{equation}
\label{eq:sae_direction}
\vv_C \;=\; \mW_{\text{dec}}\!\left( \tfrac{1}{|S_C|}\textstyle\sum_{i \in S_C} \vz_i \;-\; \tfrac{1}{|S_{\neg C}|}\textstyle\sum_{i \in S_{\neg C}} \vz_i \right),
\end{equation}
which is the mass-mean construction of \citet{marks2024geometry} adapted into the SAE feature basis, so the direction is built from the sparse features that discriminate the chapter and expressed in the coordinates the model natively computes in. As baselines we keep the raw difference-of-means direction $\vv_C^{\text{raw}}$ computed directly on $\vh$ (the strongest simple baseline in the SAE-skepticism evaluations \citep{wu2025axbench}) and a multinomial logistic probe over $\vh$ at every layer (five-fold cross-validated on the selection half), which upper-bounds what any linear read of that depth can achieve \citep{alain2016understanding}. We then use the projection $\mathrm{proj}(\vh, \vv_C) = \vh \cdot \vv_C / \lVert \vv_C \rVert$ as the basic measurement for the study.

%To build an informed foundation for clinical concept centers, we investigate eight questions raised in the subsequent section titles. % if the concept centers exist as locatable directions within model latents, if they are specifically clinical, which instrument should be used to recover them, if the model causally uses them for decision making, and if their use persists through generation.

% Move this to the method section with data and clinical concept construction
% \subsection{Which instrument should probe for clinical concept centers?}
% \label{sec:instrument}

\paragraph{Probing for concepts centers.} We compare the three most common instrumentation methods to probe the latent space of the models. The logistic probe recovers the most separability per layer, but it is a $d$-dimensional classifier with no interpretability, conflates what is encoded with what the model uses \citep{belinkov2022probing,hewitt2019designing}, and information a probe finds can be behaviorally inert \citep{elazar2021amnesic}, so we treat it as a non-interpretable calibration baseline for the other two instrumentation methods. The raw difference-of-means direction is the strongest simple carrier of a concept \citep{marks2024geometry,wu2025axbench} and needs no dictionary. The SAE direction of Equation~\ref{eq:sae_direction} is the most sophisticated, but studies have suggested that it may not be as useful \citep{kantamneni2025are}. However, as our experiments will show, the SAE turns out to be the better instrument for our setting, for three reasons. It decomposes the concept center into named, interpretable monosemantic features whose per-token firings can be monitored during generation (\S\ref{sec:temporal}, \S\ref{sec:clinician}), a single such feature nearly matches the probe ceiling (\S\ref{sec:existence}), and the SAE-derived direction is empirically the stronger causal axis than the raw direction on the same models and layers (\S\ref{sec:gensteer}). 
% We additionally study concept removal, a rank-one ablation whose controls behave differently from those of concept addition, detailed in Appendix~\ref{app:removal}.

\section{The foundational setting: Clinical concept centers}

For a reliable mechanistic analysis of models, we require two things: \textit{decodability}, meaning the concept can be read out of the latent space as a unique, distinctly separable representation \citep{alain2016understanding,park2024linear}, and \textit{causal impact}, meaning the representation demonstrably shapes the model's outputs and behaviors \citep{meng2022locating,hase2023does}.
To build an informed foundation for clinical concept centers, we investigate the questions outlined in this section as the first step in methodically establishing their existence in the latent space of LLMs.

\subsection{Do concept centers exist, and across which architectures and scales?}
\label{sec:existence}

\begin{figure}[t]
\centering
\includegraphics[width=0.44\linewidth]{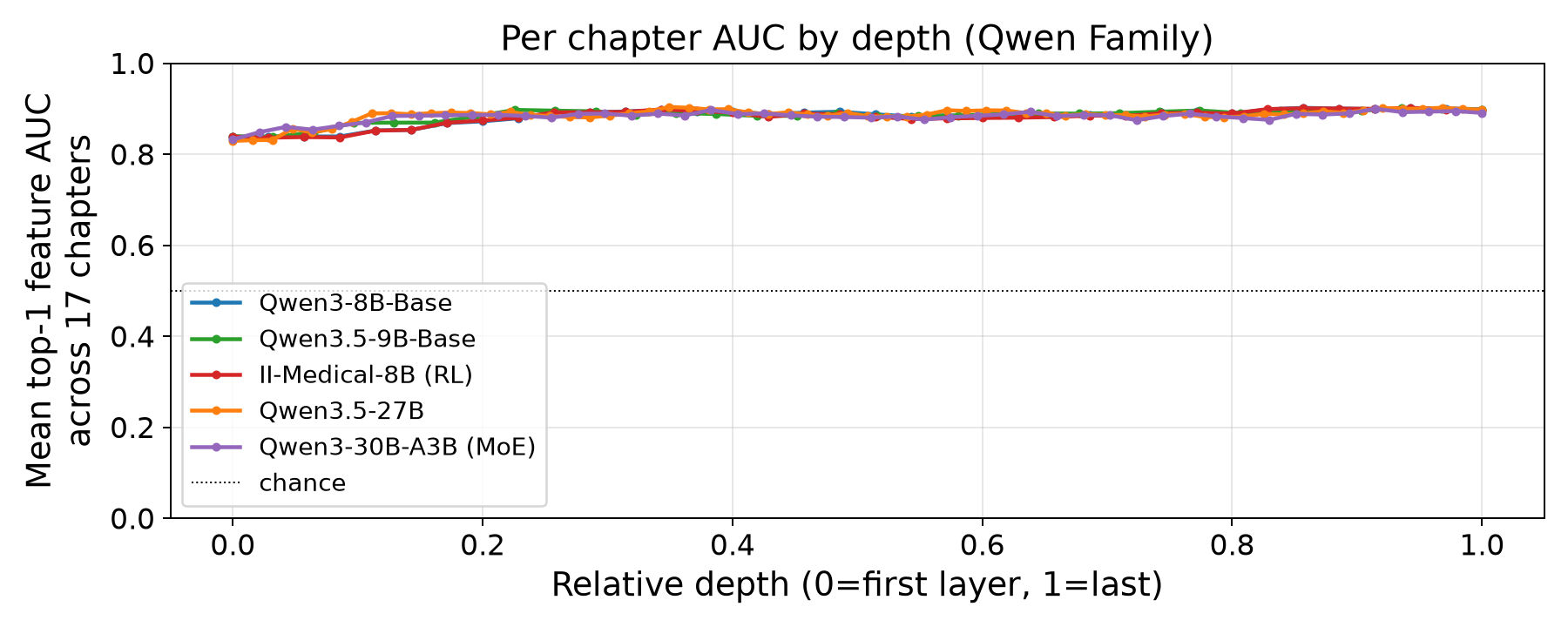}%
\hfill
\includegraphics[width=0.44\linewidth]{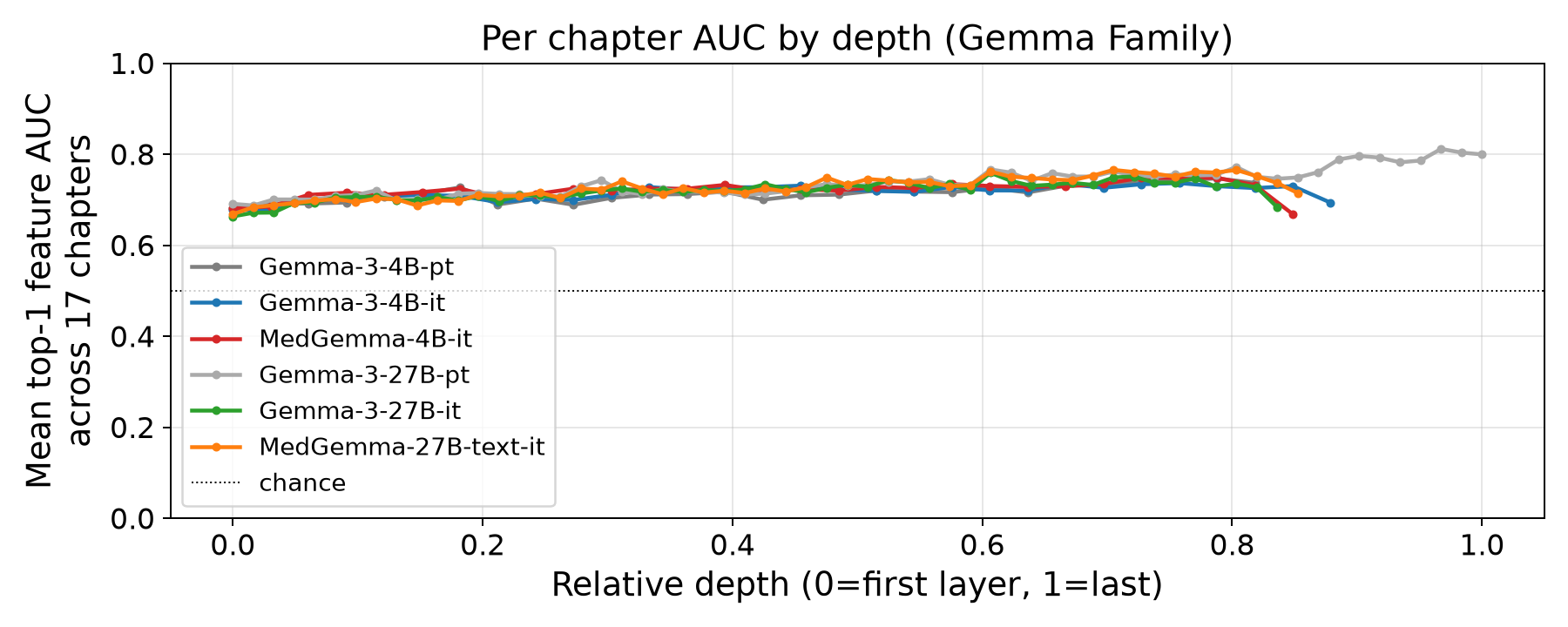}
\caption{Clinical concept centers exist across architectures and scales. One-vs-rest AUC of the best single SAE feature representing the clinical concept on the held-out set, against relative depth (left, five Qwen-family models; right, six Gemma-family models). The dashed line represents random chance. Gemma curves end before full depth where the JumpReLU dictionaries produce non-finite codes, an SAE instrument limit (only the Gemma-3-27B-pt dictionary stays finite to the last layer).}
\label{fig:auc_vs_depth}
\end{figure}

\textbf{Clinical concept centers exist in every model, architecture and scale we test}. Figure~\ref{fig:auc_vs_depth} plots the held-out one-vs-rest AUC of the best single SAE feature per chapter against depth, and the profile is the same in both families. Separability is nearly uniform across depth, starting from the the earliest layers and holding at almost the same level through the whole range the dictionaries cover. 
At its best depth, the best single feature reaches AUC 0.80--0.92 in Qwen-family and 0.70--0.90 in Gemma-family models, against a full-rank logistic-probe ceiling of 0.90--0.93 macro AUC on ten of eleven models. 
Thus, one monosemantic dictionary feature can represent most of what a strongest linear instrument (the linear probe) at that depth can find. 
Selection optimism is below 0.009 everywhere and a stricter patient-level split leaves held-out AUC above 0.877 (Appendix~\ref{app:existence}), so this separability goes beyond just in-sample selection or within-patient correlation. 
This near-uniform profile points to a consistent, reusable circuit for the chapter concept through the whole network.

These concept centers have robust properties. They are not determined by the dictionary that found them. We confirmed this with a fresh SAE trained on the medical fine-tuned II-Medical-8B (97.4\% variance explained, 0.11\% dead features), which recovers chapter directions whose residual-space cosine with the parent dictionary's is $+0.835$ on average and positive on all seventeen chapters. And they are generalizable beyond the data they were discovered with, since applied without refitting to patient-question text they retain 45--86\% of excess-over-chance separability across the Qwen family (all permutation $p < 1/2{,}000$, details provided in Appendix~\ref{app:register}). 
% which is what makes them candidates for monitoring deployed systems whose inputs do not resemble discharge summaries.

\subsection{Are the recovered features specifically clinical?}
\label{sec:specificity}

High separability could still come from cheap places like note length, lexical density, or the mere presence of disease names, so before any causal work we ask what the features respond to. To measure specificity, we take the ten highest-AUC features for each concept center and count how many activate under five conditions of decreasing clinical content: the full discharge note, the bare diagnosis text (an isolated label such as ``acute myocardial infarction''), the chapter title text (e.g., ``Diseases of the Circulatory System''), discharge notes from the adjacent chapter in the ICD-9 ordering (e.g., Respiratory notes probed with the Circulatory chapter's features), and non-clinical control sentences (short texts about geography, sports, and food). On full notes, all ten fire by design. On bare diagnosis labels, activation drops to 2.0--2.7 of ten across Qwen models, on chapter titles to 1.3--2.9, on adjacent-chapter notes to 0.3--0.6, and on non-clinical text to at most 0.016 (Appendix~\ref{app:specificity}). Thus these \textbf{features use the contextual clinical narrative (not just an isolated concept in the text), discriminate even between neighboring diagnostic categories, and do not activate outside the clinical domain}. This confirms the first requirement we set for a clinical concept center: it is specific, distinct, and unique to its clinical concept.

\subsection{Does the model causally use the concept centers for its decisions?}
\label{sec:causal}

Existence and specificity are representational facts, and representational facts can be inert \citep{elazar2021amnesic,hase2023does}, so the foundation needs a causal test. We define a seventeen-way ICD classification task as a direct test for the causal effect of these centers on the models. We ask models to classify the principal ICD chapter, with chapters mapped to answer letters (Appendix~\ref{app:prompt}). 
We add $\alpha \hat{\vv}_C$ (the chapter direction) to the residual stream at the final token position, sweep $\alpha$ over $\{\pm1,\pm3,\pm10,\pm30,\pm100\}$ and $\ell$ over every layer and select the best (layer, strength) pair. Our analysis reports only the held-out effect of this configuration, to avoid selection bias.

% Improve writing for this paragraph
% We implement four controls to isolate the causal effect. 
% % During clinical concept center finding step
% A permuted-label control rebuilds the direction with the identical construction after shuffling which notes belong to which chapter \citep{hewitt2019designing}, so if the effect survived it would never have needed the true labels. 
% % During inference step
% A wrong-chapter control injects a different chapter's direction into the same notes, so it tests whether the effect is specific to the matching concept or produced by any clinical direction. A multi-seed random control injects random unit directions under several seeds and gives the floor that any generic perturbation of the residual stream achieves. An unembedding control targets the most common shortcut explanation. The model converts its final residual state into token probabilities through the unembedding matrix, so the unembedding row of the correct answer letter is the direction that raises that letter's probability directly, and if our chapter direction worked merely by pushing the answer token instead of using the clinical concept, it would resemble and behave like this ablation. 

To attribute the steering effect to the clinical concept itself, we compare it against four controls. Each control keeps the layer, strength, held-out notes, and evaluation identical to the main experiment and alters a single component of it. Any effect a control reproduces can therefore be attributed to that component instead of the concept. The controls act at two stages of the pipeline: one alters how the concept center is \emph{discovered}, and three alter the vector \emph{injected} at inference time for steering.

\paragraph{Control at discovery time.}
Each concept center is the output of a multi-step discovery procedure: we select the features that best separate a chapter's notes from the rest (by AUC) and aggregate them into a unit direction. 
A procedure this elaborate could return a direction that improves classification even if no real concept underlies it, for instance by favoring features that are broadly active on clinical text. 
The \emph{permuted-label control} \citep{hewitt2019designing} tests this possibility. 
Before discovery, we randomly shuffle the chapter labels across notes, so that each ``chapter'' label used for feature selection becomes an arbitrary group of notes with no shared diagnosis. 
We then rerun the identical discovery procedure on these groups. 
The resulting control direction shares everything with the concept center except its grounding in real diagnoses. 
If the concept is what drives the steering effect, this direction should produce little or no improvement.

\paragraph{Controls at inference time.}
The remaining three controls take the concept centers as given and replace the steering vector added to the residual stream. 
Each asks whether something other than the matching concept could produce the same gain.
The \emph{\textbf{wrong-chapter control}} keeps the notes fixed and swaps the direction. For a note whose true chapter is $C$, we inject the direction of a different chapter $C' \neq C$ and measure the probability of $C$'s answer letter. 
If the model merely benefits from any push toward clinical content, a mismatched chapter direction should help as much as the correct one. However, if it relies on the \emph{correct} concept, the mismatched direction should fail.
The \emph{\textbf{random-direction control}} injects random unit vectors at the same layer and strength, repeated over several seeds. Adding any vector of this magnitude perturbs the model, and this control measures how much of $\Delta P$ such a generic perturbation achieves. The chapter-specific column of Table~\ref{tab:intervention} is the effect that remains after subtracting this floor.
Finally, the \emph{\textbf{unembedding control}} addresses the most common shortcut explanation for steering results. The model converts its final residual state into token probabilities through the unembedding matrix. The unembedding row of the correct answer letter is therefore the direction that most directly raises that letter's probability, without involving any clinical representation. If our chapter direction were a disguised version of this output shortcut, it would align with that row, and injecting the row would reproduce the effect.

\begin{table}[t]
\centering
\footnotesize
\setlength{\tabcolsep}{3pt}
\caption{The concept centers are causally used across both families. Every row reports $\Delta P$ (the change in correct-letter probability) during classification at the configuration selected on the selection half of the notes, evaluated on the held-out half, with bootstrap 95\% CIs (5{,}000 resamples). Chapter-specific is what remains of the injection effect after subtracting what matched-magnitude random directions achieve at the same layer and strength, thus isolating the contribution of the concept identity itself (Appendix~\ref{app:causal_detail}).}
\label{tab:intervention}

\begin{threeparttable}
\begin{tabular}{lcccccc}
\toprule
\textbf{Model} & \textbf{Layer} & \textbf{Strength} & \textbf{$\Delta P$} & \textbf{95\% CI} & \textbf{Baseline $P$} & \textbf{Chapter-specific} \\
\midrule
Qwen3-8B-Base          & 23 & +30  & +0.0685 & [+0.0652,\,+0.0719] & 0.376 & +0.044 (4 seeds) \\
Qwen3.5-9B-Base        & 19 & +10  & +0.0515 & [+0.0475,\,+0.0555] & 0.465 & +0.033 (3 seeds) \\
Qwen3-30B-A3B-Base     & 32 & +3   & +0.0692 & [+0.0645,\,+0.0739] & 0.422 & +0.094 (3 seeds) \\
Qwen3.5-27B            & 54 & +30  & +0.0406 & [+0.0364,\,+0.0447] & 0.581 & +0.057 (3 seeds) \\
II-Medical-8B          & 24 & +100 & +0.0780 & [+0.0695,\,+0.0867] & 0.111 & +0.064 (4 seeds) \\
\midrule
Gemma-3-4B-pt          & 8  & +100 & +0.0058 & [+0.0053,\,+0.0062] & 0.099 & +0.041 (3 seeds) \\
Gemma-3-4B-it          & 8  & +100 & +0.0154 & [+0.0115,\,+0.0194] & 0.443 & +0.011 (3 seeds) \\
MedGemma-4B-it\tnote{$\S$} & 6  & +100 & +0.0057 & [+0.0030,\,+0.0084] & 0.483 & $-$0.003 (3 seeds) \\
Gemma-3-27B-pt         & 5  & +100 & +0.0015 & [$-$0.0005,\,+0.0036] & 0.321 & +0.033 (3 seeds) \\
Gemma-3-27B-it         & 2  & +100 & +0.0014 & [$-$0.0005,\,+0.0036] & 0.553 & +0.000 (3 seeds) \\
MedGemma-27B-text-it   & 11 & +100 & +0.0193 & [+0.0154,\,+0.0231] & 0.496 & +0.025 (3 seeds) \\
\bottomrule
\end{tabular}

\begin{tablenotes}[flushleft]
\footnotesize
\item[$\S$] 4B dictionary is borrowed from parent model, so its results may reflect the instrument; see Appendix~\ref{app:causal_detail}.
\end{tablenotes}
\end{threeparttable}

\end{table}

Table~\ref{tab:intervention} reports our findings. The held-out \textbf{steering effect is positive on all 11 models}, with no Qwen confidence interval touching zero, and every Gemma model moving in the predicted direction, so the causal use of the concept centers holds across architectures, scales, and dictionary families (TopK and JumpReLU alike). Three properties of the effect show that it is the concept doing the work. \textbf{It is specific}, since matched-magnitude random directions achieve far less, and on the pretrained Gemma models they actively damage the model while the chapter direction helps. \textbf{It is bidirectional}, since negative strengths reduce the correct-letter probability, so the injection moves a representation the model consults rather than adding fluency. And it does not run through the output vocabulary, since the \textbf{chapter directions are essentially orthogonal to the answer letters' unembedding rows}, and injecting those rows does not reproduce the effect. The dose-response is an inverted U on every model, matching the pattern activation-steering work reports on general concepts \citep{rimsky2024steering}. Details are provided in Appendix~\ref{app:causal_detail}.

What the sweep reveals about depth matters as much as the effect itself. 
Separability is nearly uniform across depth (\S\ref{sec:existence}), yet causal effectiveness depends on the location of steering. 
The best intervention layer does not coincide with the (marginally) most separable layer on any of the five Qwen models, and instead falls in the second half of the network (roughly 60--85\% of depth, Table~\ref{tab:intervention}). 
The behavioral swing in this causally effective band is more than an order of magnitude larger than at other depths that read out the concept essentially equally well (0.283 against 0.022 on Qwen3-8B-Base, Appendix~\ref{app:causal_detail}). 
This confirms in the clinical domain the dissociation \citet{hase2023does} reported for factual editing, where localization does not predict where intervention works, and staged views of transformer depth suggest why. 
Early layers dissociate and integrate the input, the middle layers carry the complex computation over it, and the later layers govern how the model converts its latent computation into the response it presents to the user \citep{lad2024remarkable,skean2025layer,liu2025relevance}. 
The \textbf{concept centers being separable across this whole range} suggests that the models carry them through all of these functional stages, where they may support implicit computation or pass clinical information forward to later layers, even where an injection moves the output little. 
\textbf{The causal influence of an injection is largest where the model is deciding how to verbalize its latent computation}, which is why intervention there helps the most.

\subsection{Do concept centers persist through autoregressive generation?}
\label{sec:temporal}

\begin{figure}[t]
\centering
\includegraphics[width=0.98\linewidth]{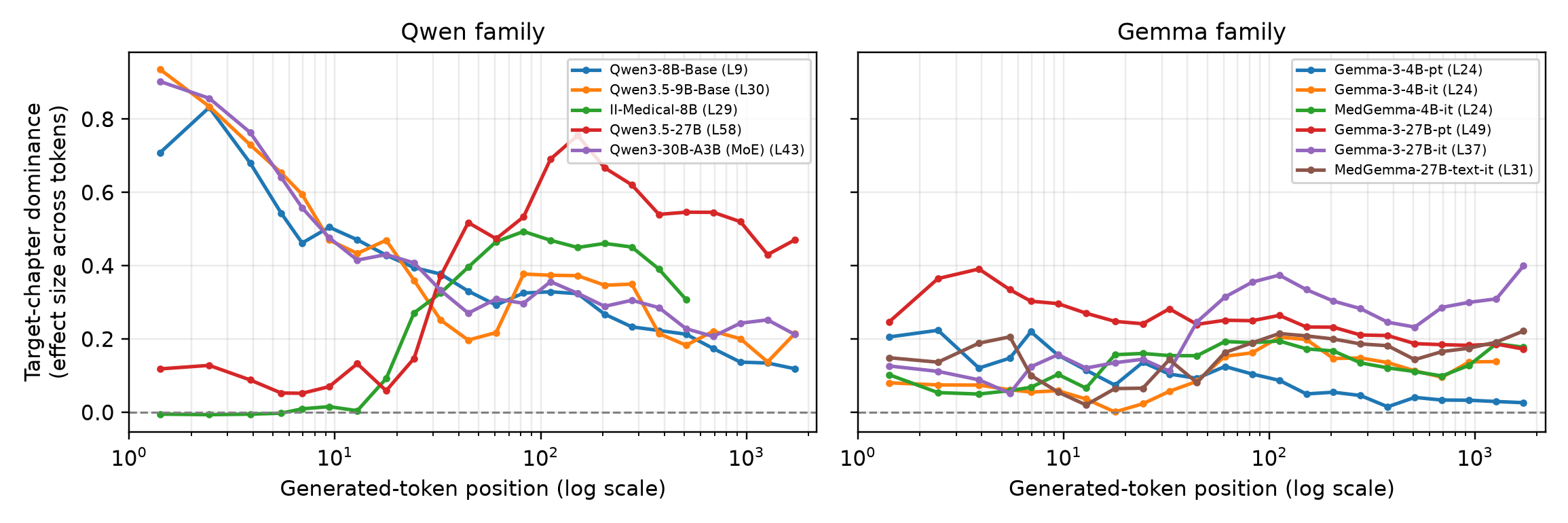}
\caption{Concept centers persist through generation. Target-concept dominance per generated token as an effect size across 170 passes per model, log-binned out to 2{,}000 tokens (measurement details in Appendix~\ref{app:temporal}).}
\label{fig:temporal}
\end{figure}

While the ICD prediction experiment establishes causal influence of the concept centers, real-world applications like monitoring a deployed system and steering an open-ended answer need the concept to stay active while text is produced. Transient circuits during token generation can be more akin to noise, and might not influence the quality of outputs during autoregressive generation. We therefore track the projection onto all 17 chapter directions at every token of long-form generation over held-out notes, 170 passes per model of up to 2{,}000 tokens, with base models continuing the note and chat models writing a prompted clinical assessment of it. Per token we reduce this to the target-chapter dominance, defined as the true chapter's projection minus the mean over all seventeen. Figure~\ref{fig:temporal} shows the dominance as an effect size across passes at each model's most representative layer. \textbf{The dominance is positive at every generated position on every model, till end of the generation window}, with Qwen3.5-27B sustaining an effect size near 0.5 from the 100th token to the 2,000th, and the Gemma family showing the same sustained shape at smaller magnitudes. The concept center is therefore a persistent state of the generation, % not a flicker at the decision token, 
which is what makes per-token monitoring meaningful. % and what we will use in the clinician evaluation in \S\ref{sec:clinician}. 
Experimental details are provided in Appendix~\ref{app:temporal}.

\section{The evaluation perspective: Counterfactual role priming}
\label{sec:roles}

With the foundation in place we can build evaluations that read the latent space instead of only the text. The first is motivated by how clinical models are deployed, since system prompts routinely assign the model a professional identity, and personas change behavior in ways output-level evaluation summarizes only poorly \citep{salewski2023incontext,zheng2024helpful}. The question a mechanistic evaluation can ask is better posed. When the model is told it is a specialist, does it move the corresponding internal clinical concept, and when the assigned role conflicts with the patient in front of it, does the model follow the role or its own internal concepts? The counterfactual design makes this measurable, presenting each note under a matched clinical specialty (the specialty of the note's true chapter), a counterfactual specialty, a randomly drawn clinical specialty, a generalist physician, a non-clinical role, and no role, and measuring both the projection onto the true chapter direction and the downstream accuracy. Projections are read at each model's best causal layer from \S\ref{sec:causal}.

\subsection{Do internal clinical concepts matter more than role based priming?}
\label{sec:roles_results}

The matched specialty raises the projection onto the matching chapter direction (paired $t = 2.9$ to $9.5$ against the counterfactual, across the ten models of both families) and converts the shift into accuracy, $+2.9$ to $+11.2$ points over the no-role baseline, while the counterfactual condition performs almost similar to the baseline (refer to Table~\ref{tab:rolepriming} in Appendix~\ref{app:roles} for per-model details). This shows that the \textbf{models are internally coherent under adversarial role priming} in this closed task. The wrong specialist role costs little to nothing because the model's own clinical concepts keep it in the right direction, while \textbf{the right specialist adds signal on top}. This pattern holds across both families, both scales, and both dictionary types.
The six-condition results further show the effect is specialty identity and not clinical framing, because a randomly drawn clinical specialty does nothing ($-0.08$ projection shift, $p = 0.83$) while matched minus random-clinical is $+1.30$ ($t = +9.57$, $p = 2.3\times10^{-19}$, 14 of 17 chapters), the generalist role sitting between (Appendix~\ref{app:roles}). 
Interestingly, this also rescues a degraded medical fine-tune whose internal chapter separation is intact while its output defaults to a single answer under constrained generation, which is further evidence for the depth picture of \S\ref{sec:causal}, since introducing the right concept signal at the causally effective depth changes what the model presents to the user (Appendix~\ref{app:roles}).

\section{The performance perspective: Model Steering}
\label{sec:steering}

With the foundation in place and the evaluation angle showing the concept centers at work, can we also leverage them to improve downstream model performance itself? In real clinical deployments there is metadata: we usually know which department the system serves, which specialty the ordering clinician belongs to, and therefore which clinical concepts are relevant before any text arrives, so a realistic intervention is to upweight the matching concept center under that prior, and investigate whether doing so improves the model answer. 
We outline the two methodological choices for steering in Appendices~\ref{app:dose} and \ref{app:sae_vs_raw}).

\paragraph{Deployment settings}\label{sec:gensteer} are open-ended and free-form, such as drafting a clinical assessment or advising on a case. We thus hold the injection active at every generated token while the model writes its own diagnosis for 1{,}624 held-out notes per model and score whether the generated text predicts the correct chapter. The result ordering aligns with our priors. \textbf{Steering along the true chapter direction helps the generation}, $+1.1$ to $+2.3$ additional correct diagnoses per hundred cases across the three models tested. \textbf{A random-direction ablation does not help, and steering with the wrong chapter hurts the model.} The holds true for all three models, spanning both families, and combines to $p = 0.00046$ (Fisher's method). Table~\ref{tab:generation_steering} in Appendix~\ref{app:dose} shows the full per-model results.

\section{Can latent concept centers predict clinician preference?}
\label{sec:clinician}

Automated evaluations like the ones above do not always capture the qualities of an output that a practicing clinician would find meaningful, so we conduct a blinded clinician evaluation to test how well our mechanistic findings align with the preferences of expert physicians. From HealthBench \citep{arora2025healthbench}, which benchmarks clinical output quality for health conversations, we filter a specialty cohort matched to the Musculoskeletal chapter. Qwen3.5-27B, the strongest chat model in our pool, answers every case under three role-priming conditions: no role, a matched-specialty role, and an adversarial other-specialty role, with the per-token SAE firings and all-layer chapter projections of every generation recorded. Two clinical raters, a senior orthopaedic resident surgeon and a junior medical student, judge the three pairwise comparisons per case blinded, position-randomized, and with a tie option, giving 228 head-to-head judgments (full protocol, exclusions, and detailed analysis in Appendix~\ref{app:clinician_protocol}). For every response we reduce the per-token concept trace at the representation-peak layer to its mean target-chapter dominance,
\begin{equation}
\mathrm{dom} \;=\; \tfrac{1}{T}\textstyle\sum_{t=1}^{T}\!\left( \mathrm{proj}_t(c^\ast) - \tfrac{1}{17}\textstyle\sum_{c} \mathrm{proj}_t(c) \right),
\end{equation}
which is the average excess activation of the cohort's chapter $c^\ast$ over the field of all chapters while the response was being written. We hypothesize that \textit{within every pair, the response that used the aligned concept center the most is the one the clinicians find more meaningful.}

\begin{table}[t]
\centering
\scriptsize
\setlength{\tabcolsep}{3pt}
\caption{Clinician preference and its agreement with activation-strength predictions.}
\label{tab:clinician_main}
\begin{tabular}{llccccc}
\toprule
& & & \multicolumn{3}{c}{\textbf{Agreement}} \\
\cmidrule(lr){4-6}
\textbf{Pairs} & \textbf{Preferred} & \textbf{Gap} & \textbf{All} & \textbf{Top half} & \textbf{Top quarter} \\
\midrule
All                  & ---          & .058 & 55.7 (.13) & 66.0 (.002) & 76.0 ($3{\times}10^{-4}$) \\
No role vs.\ adv.    & no role 50--20 & .082 & 64.3 (.022) & 77.8 (.001) & --- \\
Adv.\ vs.\ aligned   & aligned 36--24 & .057 & 53.3 (.70)  & 66.7 (.099) & --- \\
Aligned vs.\ no role & no role 41--23 & .048 & 48.4 (.90)  & 54.5 (.73)  & --- \\
\bottomrule
\end{tabular}
\end{table}

Role priming serves as a way to induce genuinely different behaviors on the same evaluation case so that the mechanistic signal has variation to predict. Table~\ref{tab:clinician_main} reports both which condition the clinicians prefer and how well the mechanistic prediction agrees with them. 
The prompt-level priming turns out to be an unreliable guide to output quality, while the mechanistic analysis aligns far more closely with the clinicians. 
The matched-specialty role is not always preferred by the clinicians in open-ended use, \textbf{so role-based priming does not guarantee better quality}. 
Within each pair, we predicted, before any rating, that the clinicians prefer the response that activated the relevant concept center more strongly while it was being generated.
Over all pairs, \textbf{as the activation strength of the aligned concept center increases, so does the agreement with clinician preference}, from 56\% over all pairs to 66\% on the half of pairs where the responses' activations separate more, and 76\% on the quarter where they separate the most. The rise appears within each comparison type, so it is the activation and not the priming condition doing the predicting, and the one comparison where agreement stays at chance (aligned versus no role) is also the one with the smallest activation gaps. 
% Thus, \textbf{when the relevant concept centers are cleanly and strongly activated, they align with clinician preference, and when they are not, the activation gap itself reveals that no reliable signal exists}. 
Thus, \textbf{the more strongly one response uses the relevant concept center than another, the more consistently clinicians prefer it}, while responses that use it to a similar degree offer no basis for a choice. 
% Together with the causal effect of steering along these centers (\S\ref{sec:gensteer}), this suggests that a model's use of its clinical concept centers bears directly on the quality of what it generates. It also points to uses beyond evaluation. In deployment settings that sample several candidate responses, the concept-center activation could serve as the selection criterion for which response to surface, and the same quantity could be reformulated as a reward signal during RL post-training. Because optimizing a measured signal can decouple it from the quality it tracks, validating these uses is left to future work.
This suggests that a mechanistic probe of the clinical concept centers can guide abstention and selection for pass-at-k behaviour in real deployment settings, and their usage in LLMs has a meaningful impact on the downstream quality of generated outputs. Notably, this also aligns with the internal coherence we find in \S\ref{sec:roles}, where linguistic role-based priming does not guide the downstream quality of the model's answers as much as its own use of the clinical concept centers.

\section{Conclusion}
\label{sec:conclusion}

We conduct a mechanistic study of clinical concept centers in LLMs. We find that they are locatable, specific, causally used by the models, persistent across generation and more effective than the linguistic substrate. We demonstrate rigorous evaluation and performance improvement of models as medical devices using these concepts, and validate it with clinician testing. Our study paves the way for future work in mechanistic interpretability of clinical AI, where the evaluation and usage of a medical model includes the latents in which it computes, not only the text in which it generates. 

% On that foundation, counterfactual role-based priming becomes a latent-space evaluation of whether models follow assigned roles or their own clinical concepts (they weigh both, measurably), and steering along the concept centers unlocks department-prior control in practical settings. We validate this with a blinded clinician study, in which the activation of the relevant concept centers predicts the clinicians' preferences exactly where the centers are cleanly and strongly activated, reaching 76\% agreement on the cases with the largest activation gaps. Extended discussion of the depth picture that runs through the results is in Appendix~\ref{app:discussion}. We see this as the necessary first step for mechanistic interpretability of clinical AI, where the evaluation of a medical model includes the space in which it computes, not only the text in which it answers.

\subsection*{Ethics statement}
This work uses MIMIC-IV discharge summaries, which are de-identified and were accessed under PhysioNet credentialed access with the required training and data use agreement. No note text, patient information, or derived data that could support re-identification appears in the paper or the released materials, and the corpus is not redistributed. The HealthBench conversations used in the clinician evaluation are public benchmark data, we generated model responses to them without any training or fine-tuning on the benchmark, and its canary string was respected. The clinician evaluation collected blinded professional judgments about machine-generated text from two clinical raters who participated voluntarily and consented to the use of their anonymized ratings and comments, and no patient data or personal information about the raters was collected. All models studied are open-weight models used within their license terms. No output of this work is intended for clinical use without regulatory review.

\subsection*{Reproducibility statement}
The supplementary material contains the code for every experiment in the paper, covering direction construction and its baselines, the specificity firing test, the injection sweep and all four controls, the role-priming conditions, the temporal traces, generation steering, and the clinician-study generation, blinding, and pre-registration pipeline, with a README describing the pipeline order and dependencies. MIMIC-IV requires PhysioNet credentialed access, and \S\ref{sec:foundation} with Appendices~\ref{app:prompt} and~\ref{app:models} describes the corpus construction, the patient-level splits, the chapter mapping, and the exact model and dictionary identifiers so the corpus can be rebuilt from the source data. The held-out selection protocol, seeds, and statistical conventions for the causal claims are given in \S\ref{sec:causal} and Appendix~\ref{app:causal_detail}, the full clinician-study protocol including the sealed answer key, the frozen pre-registered predictions, and the capability-exclusion list is in Appendix~\ref{app:clinician_protocol}, and the curated rater workbooks are released with the data so every reported statistic in that section can be recomputed from the same filtered records.

\subsection*{AI use statement}
We used an LLM-based coding assistant, under continuous author direction, for tasks in both the required and recommended disclosure categories. In the required category, the assistant implemented parts of the experiment and evaluation code, executed the experiment sweeps on our hardware, performed the statistical computations reported in the paper and reformatted the MIMIC-IV notes into the labeled corpus described in \S\ref{sec:foundation}. All generated code was reviewed by the authors. We have not used generative AI to generate synthetic datasets, to formulate or prove mathematical claims, or to propose the hypotheses under test. We reviewed all AI-assisted work. All experiment code was reviewed by the authors and is included in the supplementary material, every reported number was traced to a stored experiment artifact, all citations were verified against the cited papers, and the clinician study was designed, pre-registered, and adjudicated by the authors. We take responsibility for the final content of this work, including text, claims, and artifacts produced with the aid of generative AI.

\bibliography{references}

@article{templeton2026scaling,
  title={Scaling monosemanticity: Extracting interpretable features from claude 3 sonnet},
  author={Templeton, Adly and Conerly, Tom and Marcus, Jonathan and Lindsey, Jack and Bricken, Trenton and Chen, Brian and Pearce, Adam and Citro, Craig and Ameisen, Emmanuel and Jones, Andy and others},
  journal={arXiv preprint arXiv:2605.29358},
  year={2026}
}

@inproceedings{cunningham2024sparse,
  title={Sparse Autoencoders Find Highly Interpretable Features in Language Models},
  author={Cunningham, Hoagy and Ewart, Aidan and Riggs, Logan and Huben, Robert and Sharkey, Lee},
  booktitle={The Twelfth International Conference on Learning Representations (ICLR)},
  year={2024}
}

@article{rajamanoharan2024jumping,
  title={Jumping Ahead: Improving Reconstruction Fidelity with {JumpReLU} Sparse Autoencoders},
  author={Rajamanoharan, Senthooran and Lieberum, Tom and Sonnerat, Nicolas and Conmy, Arthur and Varma, Vikrant and Kram{\'a}r, J{\'a}nos and Shah, Rohin},
  journal={arXiv preprint arXiv:2407.14435},
  year={2024}
}

@article{deng2026qwen,
  title={Qwen-scope: Turning sparse features into development tools for large language models},
  author={Deng, Boyi and Wang, Xu and Wang, Yaoning and Wan, Yu and Ma, Yubo and Yang, Baosong and Wei, Haoran and Tang, Jialong and Lin, Huan and Gao, Ruize and others},
  journal={arXiv preprint arXiv:2605.11887},
  year={2026}
}

@article{kantamneni2025are,
  title={Are Sparse Autoencoders Useful? A Case Study in Sparse Probing},
  author={Kantamneni, Subhash and Engels, Joshua and Rajamanoharan, Senthooran and Tegmark, Max and Nanda, Neel},
  journal={arXiv preprint arXiv:2502.16681},
  year={2025}
}

@inproceedings{wu2025axbench,
  title={{AxBench}: Steering {LLM}s? Even Simple Baselines Outperform Sparse Autoencoders},
  author={Wu, Zhengxuan and Arora, Aryaman and Geiger, Atticus and Wang, Zheng and Huang, Jing and Jurafsky, Dan and Manning, Christopher D. and Potts, Christopher},
  booktitle={Proceedings of the 42nd International Conference on Machine Learning (ICML)},
  year={2025}
}

@inproceedings{karvonen2025saebench,
  title={{SAEBench}: A Comprehensive Benchmark for Sparse Autoencoders in Language Model Interpretability},
  author={Karvonen, Adam and Rager, Can and Lin, Johnny and Tigges, Curt and Bloom, Joseph and Chanin, David and Lau, Yeu-Tong and Farrell, Eoin and Conmy, Arthur and McDougall, Callum and others},
  booktitle={Proceedings of the 42nd International Conference on Machine Learning (ICML)},
  year={2025}
}

@article{chanin2024absorption,
  title={A is for Absorption: Studying Feature Splitting and Absorption in Sparse Autoencoders},
  author={Chanin, David and Wilken-Smith, James and Dulka, Tom{\'a}{\v{s}} and Bhatnagar, Hardik and Bloom, Joseph},
  journal={arXiv preprint arXiv:2409.14507},
  year={2024}
}

@article{engels2024darkmatter,
  title={Decomposing the Dark Matter of Sparse Autoencoders},
  author={Engels, Joshua and Riggs, Logan and Tegmark, Max},
  journal={arXiv preprint arXiv:2410.14670},
  year={2024}
}

@article{paulo2025sparse,
  title={Sparse Autoencoders Trained on the Same Data Learn Different Features},
  author={Paulo, Gon{\c{c}}alo and Belrose, Nora},
  journal={arXiv preprint arXiv:2501.16615},
  year={2025}
}

@article{karvonen2024evaluating,
  title={Evaluating sparse autoencoders on targeted concept erasure tasks},
  author={Karvonen, Adam and Rager, Can and Marks, Samuel and Nanda, Neel and others},
  journal={arXiv preprint arXiv:2411.18895},
  year={2024}
}

@inproceedings{rimsky2024steering,
  title={Steering Llama 2 via Contrastive Activation Addition},
  author={Rimsky, Nina and Gabrieli, Nick and Schulz, Julian and Tong, Meg and Hubinger, Evan and Turner, Alexander Matt},
  booktitle={Proceedings of the 62nd Annual Meeting of the Association for Computational Linguistics (ACL)},
  pages={15504--15522},
  year={2024}
}

@inproceedings{turner2024steering,
  title={Steering Language Models with Activation Engineering},
  author={Turner, Alexander Matt and Thiergart, Lisa and Leech, Gavin and Udell, David and Vazquez, Juan J. and Mini, Ulisse and MacDiarmid, Monte},
  booktitle={arXiv preprint arXiv:2308.10248},
  year={2023}
}

@article{zou2023representation,
  title={Representation Engineering: A Top-Down Approach to {AI} Transparency},
  author={Zou, Andy and Phan, Long and Chen, Sarah and Campbell, James and Guo, Phillip and Ren, Richard and Pan, Alexander and Yin, Xuwang and Mazeika, Mantas and Dombrowski, Ann-Kathrin and others},
  journal={arXiv preprint arXiv:2310.01405},
  year={2023}
}

@inproceedings{tan2024analysing,
  title={Analysing the Generalisation and Reliability of Steering Vectors},
  author={Tan, Daniel and Chanin, David and Lynch, Aengus and Paige, Brooks and Kanoulas, Dimitrios and Garriga-Alonso, Adri{\`a} and Kirk, Robert},
  booktitle={Advances in Neural Information Processing Systems (NeurIPS)},
  pages={139179--139212},
  year={2024}
}

@inproceedings{belrose2023leace,
  title={{LEACE}: Perfect Linear Concept Erasure in Closed Form},
  author={Belrose, Nora and Schneider-Joseph, David and Ravfogel, Shauli and Cotterell, Ryan and Raff, Edward and Biderman, Stella},
  booktitle={Advances in Neural Information Processing Systems (NeurIPS)},
  year={2023}
}

@inproceedings{hewitt2019designing,
  title={Designing and interpreting probes with control tasks},
  author={Hewitt, John and Liang, Percy},
  booktitle={Proceedings of the 2019 Conference on Empirical Methods in Natural Language Processing and the 9th International Joint Conference on Natural Language Processing (EMNLP-IJCNLP)},
  pages={2733--2743},
  year={2019}
}

@article{elazar2021amnesic,
  title={Amnesic probing: Behavioral explanation with amnesic counterfactuals},
  author={Elazar, Yanai and Ravfogel, Shauli and Jacovi, Alon and Goldberg, Yoav},
  journal={Transactions of the Association for Computational Linguistics},
  volume={9},
  pages={160--175},
  year={2021}
}

@article{marks2025sparse,
  title={Sparse Feature Circuits: Discovering and Editing Interpretable Causal Graphs in Language Models},
  author={Marks, Samuel and Rager, Can and Michaud, Eric J. and Belinkov, Yonatan and Bau, David and Mueller, Aaron},
  journal={The Thirteenth International Conference on Learning Representations (ICLR)},
  year={2025}
}

@inproceedings{movva2025sparse,
  title={Sparse autoencoders for hypothesis generation},
  author={Movva, Rajiv and Peng, Kenny and Garg, Nikhil and Kleinberg, Jon and Pierson, Emma},
  booktitle={Proceedings of the 42nd International Conference on Machine Learning (ICML)},
  year={2025}
}

@inproceedings{gallifant2025sparse,
  title={Sparse autoencoder features for classifications and transferability},
  author={Gallifant, Jack and Chen, Shan and Sasse, Kuleen and Aerts, Hugo and Hartvigsen, Thomas and Bitterman, Danielle S.},
  booktitle={Proceedings of the 2025 Conference on Empirical Methods in Natural Language Processing (EMNLP)},
  year={2025}
}

@inproceedings{hase2023does,
  title={Does Localization Inform Editing? Surprising Differences in Causality-Based Localization vs. Knowledge Editing in Language Models},
  author={Hase, Peter and Bansal, Mohit and Kim, Been and Ghandeharioun, Asma},
  booktitle={Advances in Neural Information Processing Systems (NeurIPS)},
  year={2023}
}

@article{lad2024remarkable,
  title={The Remarkable Robustness of {LLM}s: Stages of Inference?},
  author={Vedang Lad and Jin Hwa Lee and Wes Gurnee and Max Tegmark},
  journal={arXiv preprint arXiv:2406.19384},
  year={2024}
}

@inproceedings{skean2025layer,
  title={Layer by Layer: Uncovering Hidden Representations in Language Models},
  author={Skean, Oscar and Arefin, Md Rifat and Zhao, Dan and Patel, Niket and Naghiyev, Jalal and LeCun, Yann and Shwartz-Ziv, Ravid},
  booktitle={Proceedings of the 42nd International Conference on Machine Learning (ICML)},
  year={2025}
}

@inproceedings{meng2022locating,
  title={Locating and Editing Factual Associations in {GPT}},
  author={Meng, Kevin and Bau, David and Andonian, Alex and Belinkov, Yonatan},
  booktitle={Advances in Neural Information Processing Systems (NeurIPS)},
  year={2022}
}

@article{PhysioNet-mimic-iv-note-2.2,
  author = {Johnson, Alistair and Pollard, Tom and Horng, Steven and Celi, Leo Anthony and Mark, Roger},
  title = {{MIMIC-IV-Note: Deidentified free-text clinical notes}},
  journal = {{PhysioNet}},
  year = {2023},
  month = jan,
  note = {Version 2.2},
  doi = {10.13026/1n74-ne17},
  url = {https://doi.org/10.13026/1n74-ne17}
}

@inproceedings{mullenbach2018explainable,
  title={Explainable Prediction of Medical Codes from Clinical Text},
  author={Mullenbach, James and Wiegreffe, Sarah and Duke, Jon and Sun, Jimeng and Eisenstein, Jacob},
  booktitle={Proceedings of the 2018 Conference of the North American Chapter of the Association for Computational Linguistics (NAACL-HLT)},
  pages={1101--1111},
  year={2018}
}

@inproceedings{kweon2024ehrnoteqa,
  title={EHRNoteQA: An LLM Benchmark for Real-World Clinical Practice Using Discharge Summaries},
  author={Kweon, Sunjun and Kim, Jiyoun and Kwak, Heeyoung and Cha, Dongchul and Yoon, Hangyul and Kim, Kwanhyung and Yang, Jeewon and Won, Seunghyun and Choi, Edward},
  booktitle={Advances in Neural Information Processing Systems (NeurIPS) Datasets and Benchmarks Track},
  volume={37},
  year={2024}
}

@article{orgad2025llms,
  title={{LLM}s Know More Than They Show: On the Intrinsic Representation of {LLM} Hallucinations},
  author={Orgad, Hadas and Toker, Michael and Gekhman, Zorik and Reichart, Roi and Szpektor, Idan and Kotek, Hadas and Belinkov, Yonatan},
  journal={The Thirteenth International Conference on Learning Representations (ICLR)},
  year={2025}
}

@article{sellergren2025medgemma,
  title={{MedGemma} Technical Report},
  author={Sellergren, Andrew and Kazemzadeh, Sahar and Jaroensri, Tiam and Kiraly, Atilla and Traverse, Madeleine and Kohlberger, Timo and Xu, Shawn and Jamil, Fayaz and Hughes, C{\'\i}an and Lau, Charles and others},
  journal={arXiv preprint arXiv:2507.05201},
  year={2025}
}

@article{singhal2023large,
  title={Large Language Models Encode Clinical Knowledge},
  author={Singhal, Karan and Azizi, Shekoofeh and Tu, Tao and Mahdavi, S. Sara and Wei, Jason and Chung, Hyung Won and Scales, Nathan and Tanwani, Ajay and Cole-Lewis, Heather and Pfohl, Stephen and others},
  journal={Nature},
  volume={620},
  number={7972},
  pages={172--180},
  year={2023},
  publisher={Nature Publishing Group}
}

@article{singhal2025toward,
  title={Toward Expert-Level Medical Question Answering with Large Language Models},
  author={Singhal, Karan and Tu, Tao and Gottweis, Juraj and Sayres, Rory and Wulczyn, Ellery and Amin, Mohamed and Hou, Le and Clark, Kevin and Pfohl, Stephen R. and Cole-Lewis, Heather and others},
  journal={Nature Medicine},
  volume={31},
  pages={943--950},
  year={2025},
  publisher={Nature Publishing Group}
}

@article{goh2024large,
  title={Large Language Model Influence on Diagnostic Reasoning: A Randomized Clinical Trial},
  author={Goh, Ethan and Gallo, Robert and Hom, Jason and Strong, Eric and Weng, Yingjie and Kerman, Hannah and Cool, Joseph A. and Kanjee, Zahir and Parsons, Andrew S. and Ahuja, Neera and others},
  journal={JAMA Network Open},
  volume={7},
  number={10},
  pages={e2440969},
  year={2024},
  publisher={American Medical Association}
}

@article{hager2024evaluation,
  title={Evaluation and Mitigation of the Limitations of Large Language Models in Clinical Decision-Making},
  author={Hager, Paul and Jungmann, Friederike and Holland, Robbie and Bhagat, Kunal and Hubrecht, Inga and Knauer, Manuel and Vielhauer, Jakob and Makowski, Marcus and Braren, Rickmer and Kaissis, Georgios and Rueckert, Daniel},
  journal={Nature Medicine},
  volume={30},
  number={9},
  pages={2613--2622},
  year={2024},
  publisher={Nature Publishing Group}
}

@article{arora2025healthbench,
  title={Healthbench: Evaluating large language models towards improved human health},
  author={Arora, Rahul K and Wei, Jason and Hicks, Rebecca Soskin and Bowman, Preston and Qui{\~n}onero-Candela, Joaquin and Tsimpourlas, Foivos and Sharman, Michael and Shah, Meghan and Vallone, Andrea and Beutel, Alex and others},
  journal={arXiv preprint arXiv:2505.08775},
  year={2025}
}

@article{jin2021disease,
  title={What Disease Does This Patient Have? A Large-Scale Open Domain Question Answering Dataset from Medical Exams},
  author={Jin, Di and Pan, Eileen and Oufattole, Nassim and Weng, Wei-Hung and Fang, Hanyi and Szolovits, Peter},
  journal={Applied Sciences},
  volume={11},
  number={14},
  pages={6421},
  year={2021},
  publisher={MDPI}
}

@inproceedings{pal2022medmcqa,
  title={{MedMCQA}: A Large-Scale Multi-Subject Multi-Choice Dataset for Medical Domain Question Answering},
  author={Pal, Ankit and Umapathi, Logesh Kumar and Sankarasubbu, Malaikannan},
  booktitle={Proceedings of the Conference on Health, Inference, and Learning (CHIL)},
  pages={248--260},
  year={2022}
}

@inproceedings{turpin2023language,
  title={Language Models Don't Always Say What They Think: Unfaithful Explanations in Chain-of-Thought Prompting},
  author={Turpin, Miles and Michael, Julian and Perez, Ethan and Bowman, Samuel R.},
  booktitle={Advances in Neural Information Processing Systems (NeurIPS)},
  year={2023}
}

@article{lanham2023measuring,
  title={Measuring Faithfulness in Chain-of-Thought Reasoning},
  author={Lanham, Tamera and Chen, Anna and Radhakrishnan, Ansh and Steiner, Benoit and Denison, Carson and Hernandez, Danny and Li, Dustin and Durmus, Esin and Hubinger, Evan and Kernion, Jackson and others},
  journal={arXiv preprint arXiv:2307.13702},
  year={2023}
}

@article{abdulaal2024interpretable,
  title={An {X}-{R}ay Is Worth 15 Features: Sparse Autoencoders for Interpretable Radiology Report Generation},
  author={Abdulaal, Ahmed and Fry, Hugo and Mont{\~n}a-Brown, Nina and Ijishakin, Ayodeji and Gao, Jack and Hyland, Stephanie and Alexander, Daniel C. and Castro, Daniel C.},
  journal={arXiv preprint arXiv:2410.03334},
  year={2024}
}

@article{o2025resurrecting,
  title={Resurrecting the salmon: Rethinking mechanistic interpretability with domain-specific sparse autoencoders},
  author={O'Neill, Charles and Jayasekara, Mudith and Kirkby, Max},
  journal={arXiv preprint arXiv:2508.09363},
  year={2025}
}

@misc{modi2026clinical,
  author={Modi, Mirage and Krull, Jordan E. and Johnson, David and Wang, Xinyu and Gauntner, Timothy D. and Li, Ming and Cheng, Hao and Ma, Anjun and Zhang, Ping and Stover, Daniel G. and Li, Zihai and Ma, Qin},
  title={Understanding Clinical Reasoning Variability in Medical Large Language Models: A Mechanistic Interpretability Study},
  year={2026},
  howpublished={medRxiv 2026.01.26.26344845}
}

@article{navarro2026internal,
  title={Internal Representation, Not Clinical Knowledge: Where Apparent {LLM} Triage Failures Originate},
  author={Fraile Navarro, David and Como, Berardino and Sheng, Jialei and Ananthan, Soundariya},
  journal={arXiv preprint arXiv:2605.29889},
  year={2026}
}

@inproceedings{park2024linear,
  title={The Linear Representation Hypothesis and the Geometry of Large Language Models},
  author={Park, Kiho and Choe, Yo Joong and Veitch, Victor},
  booktitle={Proceedings of the 41st International Conference on Machine Learning (ICML)},
  year={2024}
}

@inproceedings{marks2024geometry,
  title={The Geometry of Truth: Emergent Linear Structure in Large Language Model Representations of True/False Datasets},
  author={Marks, Samuel and Tegmark, Max},
  booktitle={Conference on Language Modeling (COLM)},
  year={2024}
}

@article{alain2016understanding,
  title={Understanding Intermediate Layers Using Linear Classifier Probes},
  author={Alain, Guillaume and Bengio, Yoshua},
  journal={arXiv preprint arXiv:1610.01644},
  year={2016}
}

@article{belinkov2022probing,
  title={Probing Classifiers: Promises, Shortcomings, and Advances},
  author={Belinkov, Yonatan},
  journal={Computational Linguistics},
  volume={48},
  number={1},
  pages={207--219},
  year={2022},
  publisher={MIT Press}
}

@inproceedings{yue2025does,
  title={Does Reinforcement Learning Really Incentivize Reasoning Capacity in {LLMs} Beyond the Base Model?},
  author={Yue, Yang and Chen, Zhiqi and Lu, Rui and Zhao, Andrew and Wang, Zhaokai and Song, Shiji and Huang, Gao},
  booktitle={Advances in Neural Information Processing Systems (NeurIPS)},
  year={2025},
  note={arXiv:2504.13837}
}

@article{wang2025resa,
  title={{Resa}: Transparent Reasoning Models via {SAE}s},
  author={Wang, Shangshang and Asilis, Julian and Akg{\"u}l, {\"O}mer Faruk and Bilgin, Enes Burak and Liu, Ollie and Fu, Deqing and Neiswanger, Willie},
  journal={arXiv preprint arXiv:2506.09967},
  year={2025}
}

@article{johnson2023mimic,
  title={{MIMIC-IV}, a Freely Accessible Electronic Health Record Dataset},
  author={Johnson, Alistair E. W. and Bulgarelli, Lucas and Shen, Lu and Gayles, Alvin and Shammout, Ayad and Horng, Steven and Pollard, Tom J. and Hao, Sicheng and Moody, Benjamin and Gow, Brian and others},
  journal={Scientific Data},
  volume={10},
  number={1},
  pages={1},
  year={2023},
  publisher={Nature Publishing Group}
}

@inproceedings{zheng2024helpful,
  title={When ``A Helpful Assistant'' Is Not Really Helpful: Personas in System Prompts Do Not Improve Performances of Large Language Models},
  author={Zheng, Mingqian and Pei, Jiaxin and Logeswaran, Lajanugen and Lee, Moontae and Jurgens, David},
  booktitle={Findings of the Association for Computational Linguistics: EMNLP 2024},
  year={2024}
}

@inproceedings{salewski2023incontext,
  title={In-Context Impersonation Reveals Large Language Models' Strengths and Biases},
  author={Salewski, Leonard and Alaniz, Stephan and Rio-Torto, Isabel and Schulz, Eric and Akata, Zeynep},
  booktitle={Advances in Neural Information Processing Systems (NeurIPS)},
  year={2023}
}

@inproceedings{casper2024black,
  title={Black-Box Access Is Insufficient for Rigorous {AI} Audits},
  author={Casper, Stephen and Ezell, Carson and Siegmann, Charlotte and Kolt, Noam and Curtis, Taylor Lynn and Bucknall, Benjamin and Haupt, Andreas and Wei, Kevin and Scheurer, J{\'e}r{\'e}my and Hobbhahn, Marius and others},
  booktitle={Proceedings of the 2024 ACM Conference on Fairness, Accountability, and Transparency (FAccT)},
  pages={2254--2272},
  year={2024}
}

@article{mesko2023imperative,
  title={The Imperative for Regulatory Oversight of Large Language Models (or Generative {AI}) in Healthcare},
  author={Mesk{\'o}, Bertalan and Topol, Eric J.},
  journal={npj Digital Medicine},
  volume={6},
  number={1},
  pages={120},
  year={2023},
  publisher={Nature Publishing Group}
}

@inproceedings{paszke2019pytorch,
  title={{PyTorch}: An Imperative Style, High-Performance Deep Learning Library},
  author={Paszke, Adam and Gross, Sam and Massa, Francisco and Lerer, Adam and Bradbury, James and Chanan, Gregory and Killeen, Trevor and Lin, Zeming and Gimelshein, Natalia and Antiga, Luca and others},
  booktitle={Advances in Neural Information Processing Systems (NeurIPS)},
  volume={32},
  year={2019}
}

@inproceedings{wolf2020transformers,
  title={Transformers: State-of-the-Art Natural Language Processing},
  author={Wolf, Thomas and Debut, Lysandre and Sanh, Victor and Chaumond, Julien and Delangue, Clement and Moi, Anthony and Cistac, Pierric and Rault, Tim and Louf, R{\'e}mi and Funtowicz, Morgan and others},
  booktitle={Proceedings of the 2020 Conference on Empirical Methods in Natural Language Processing: System Demonstrations},
  pages={38--45},
  year={2020}
}

@misc{bloom2024saetrainingcodebase,
   title = {SAELens},
   author = {Bloom, Joseph and Tigges, Curt and Duong, Anthony and Chanin, David},
   year = {2024},
   howpublished = {\url{https://github.com/decoderesearch/SAELens}},
}

@misc{nanda2022transformerlens,
    title = {TransformerLens},
    author = {Neel Nanda and Joseph Bloom},
    year = {2022},
    howpublished = {\url{https://github.com/TransformerLensOrg/TransformerLens}},
}

@article{liu2025relevance,
  title={How Do Large Language Models Understand Relevance? A Mechanistic Interpretability Perspective},
  author={Liu, Qi and Duan, Haozhe and Mao, Jiaxin and Wen, Ji-Rong},
  journal={ACM Transactions on Information Systems},
  volume={44},
  number={2},
  pages={34:1--34:27},
  year={2025},
  publisher={ACM}
}

@article{nagar2026clrvoyance,
  title={{CLR}-voyance: Reinforcing Open-Ended Reasoning for Inpatient Clinical Decision Support with Outcome-Aware Rubrics},
  author={Nagar, Aishik and Kaliya-Perumal, Arun-Kumar and Han, Yu-Hsuan and Huang, Andrew Sheng-Han and Kee, Kristen and Cao, Yushi and Chen, Yiming and Jiang, Hongchao},
  journal={arXiv preprint arXiv:2605.09584},
  year={2026}
}

@article{singh2023explaining,
  title={Explaining Black Box Text Modules in Natural Language with Language Models},
  author={Singh, Chandan and Hsu, Aliyah R. and Antonello, Richard and Jain, Shailee and Huth, Alexander G. and Yu, Bin and Gao, Jianfeng},
  journal={arXiv preprint arXiv:2305.09863},
  year={2023}
}

@article{paulo2024automatically,
  title={Automatically Interpreting Millions of Features in Large Language Models},
  author={Paulo, Gon{\c{c}}alo and Mallen, Alex and Juang, Caden and Belrose, Nora},
  journal={arXiv preprint arXiv:2410.13928},
  year={2024}
}

@inproceedings{choi2017gram,
  title={{GRAM}: Graph-Based Attention Model for Healthcare Representation Learning},
  author={Choi, Edward and Bahadori, Mohammad Taha and Song, Le and Stewart, Walter F. and Sun, Jimeng},
  booktitle={Proceedings of the 23rd ACM SIGKDD International Conference on Knowledge Discovery and Data Mining (KDD)},
  pages={787--795},
  year={2017}
}

@article{mamalakis2026monosemantic,
  title={A Monosemantic Attribution Framework for Stable Interpretability in Clinical Neuroscience Transformer-Based Language Models},
  author={Mamalakis, Michail and Azevedo, Tiago and Cosentino, Cristian and D'Ercoli, Chiara and Abulikemu, Subati and Sun, Zhongtian and Bethlehem, Richard and Lio, Pietro},
  journal={arXiv preprint arXiv:2601.17952},
  year={2026}
}

@article{lehnschioler2026eeg,
  title={Mechanistic Interpretability of {EEG} Foundation Models via Sparse Autoencoders},
  author={Lehn-Schi{\o}ler, William and Kj{\ae}r, Magnus Ruud and Thapa, Rahul and Pedersen, Magnus Guldberg and Storgaard, Anton Mosquera and Williams, Nick and G{\u a}tej, Radu and Lehn-Schi{\o}ler, Tue and Brink-Kj{\ae}r, Andreas and Puthusserypady, Sadasivan and Beniczky, S{\'a}ndor and Zou, James and Hansen, Lars Kai},
  journal={arXiv preprint arXiv:2605.13930},
  year={2026}
}
\bibliographystyle{iclr2027_conference}

\appendix

\section{Extended related work}
\label{app:related}

This appendix extends \S\ref{sec:related} with the lines of work that further shaped our controls and design choices.

\paragraph{SAE features as practical instruments.}
\citet{gallifant2025sparse} evaluate SAE features as classifier inputs for safety-critical text classification and find them competitive and transferable across models and languages, which is the supervised regime where we, consistent with \citet{kantamneni2025are}, also expect no SAE advantage over a full-rank probe. \citet{movva2025sparse} use SAE features over text embeddings to generate natural-language hypotheses that predict labels, the closest use to ours in spirit, since both treat the dictionary as a discovery instrument over a labeled corpus. Our formulation differs by grounding the labels in an expert-maintained clinical ontology rather than generating them, and by testing the recovered representations causally rather than correlationally.

\paragraph{Known caveats of sparse dictionaries.}
Sparse dictionaries absorb sub-concepts into broader features \citep{chanin2024absorption}, leave a share of the signal unexplained \citep{engels2024darkmatter}, learn different features when retrained on the same data \citep{paulo2025sparse}, and are themselves evaluated through targeted concept-erasure tasks \citep{karvonen2024evaluating}. These caveats motivated three of our design choices, the coarse chapter-level concept granularity (absorption pressure is highest for fine-grained concepts), the dense all-layer raw-residual projections kept alongside every SAE trace (so signal the dictionary misses is not lost), and the dictionary-independence check of \S\ref{sec:existence}, where a freshly trained dictionary recovers the same chapter directions as the borrowed parent one.

\paragraph{Reliability of steering vectors.}
\citet{tan2024analysing} show that steering vectors generalize unreliably, with high variance across inputs and settings, which motivated our relative-dose rule, the mandatory discordance counts before any null is believed (Appendix~\ref{app:dose}), and the multi-seed random controls reported next to every intervention effect. \citet{marks2025sparse} scale circuit-level analysis with SAE features, which is the natural next granularity below our chapter-level directions and the instrument we expect to need when moving from chapters to finer clinical concepts.

\paragraph{Interpretability of general purpose LLMs.}
Recent applications of sparse autoencoders (SAEs) in the clinical domain have largely been used for interpretability studies for specific architectures or specific clinical sub-domains \citep{mamalakis2026monosemantic, lehnschioler2026eeg}. While these studies successfully decompose the models across these specialised settings, their findings do not readily generalise to broader foundation models. We bridge this generalization gap by investigating general purpose large language models, specifically the Qwen and Gemma families, demonstrating that clinically relevant concept centers natively emerge within their latent spaces. Furthermore, we establish the practical utility of these concept centers by leveraging them to improve ICD-9 chapter classification accuracy and align model outputs with clinician preferences.

\section{Compute and toolkit used}
Our experiments run on PyTorch \citep{paszke2019pytorch} over HuggingFace Transformers checkpoints \citep{wolf2020transformers}, with Gemma dictionaries loaded through SAELens \citep{bloom2024saetrainingcodebase} and hook-level activation access in the style of TransformerLens \citep{nanda2022transformerlens}. Experiments were carried out on a combination of Nvidia H100 and A100 GPUs.

\section{Prompt template and chapter mapping}
\label{app:prompt}

This appendix fixes the classification task used throughout \S\ref{sec:causal} and \S\ref{sec:roles}. The prompt template is shown below. The discharge summary text includes all sections as they appear in MIMIC-IV-Note, including the Discharge Diagnosis and Discharge Condition sections, with no truncation or filtering of the clinical narrative.

\begin{verbatim}
You are a clinical coding assistant. Read the discharge
summary below and identify which ICD-9 disease chapter
the principal diagnosis belongs to.

Discharge Summary:
{NOTE_TEXT}

Question: Which of the following ICD-9 chapters does the
principal diagnosis fall under?

A. Infectious and Parasitic Diseases (001-139)
B. Neoplasms (140-239)
C. Endocrine, Nutritional, Metabolic, and Immunity (240-279)
D. Diseases of Blood and Blood-Forming Organs (280-289)
E. Mental Disorders (290-319)
F. Diseases of the Nervous System (320-389)
G. Diseases of the Sense Organs (390-459)
H. Diseases of the Circulatory System (460-519)
I. Diseases of the Respiratory System (520-579)
J. Diseases of the Digestive System (580-629)
K. Diseases of the Genitourinary System (630-679)
L. Complications of Pregnancy, Childbirth, Puerperium (680-709)
M. Diseases of the Skin and Subcutaneous Tissue (710-739)
N. Diseases of the Musculoskeletal System (740-759)
O. Congenital Anomalies (760-779)
P. Certain Conditions Originating in Perinatal Period (780-799)
Q. Symptoms, Signs, and Ill-Defined Conditions (800-999)

Answer (letter only):
\end{verbatim}

\textbf{Decoding parameters.} Temperature 0.0, top-p 1.0, maximum 10 new tokens. Context length is 8{,}192 tokens for all models except Gemma-3-27B and Qwen3.5-27B (32{,}768). Notes exceeding the context length are truncated from the end.

Table~\ref{tab:icd9-chapters} lists the seventeen ICD-9-CM disease chapters, their code ranges, and per-chapter counts in the full 8{,}617-note manifest after per-chapter capping at 1{,}000.

\begin{table}[h]
\centering
\small
\begin{tabular}{llrr}
\hline
\textbf{Letter} & \textbf{Chapter name} & \textbf{ICD-9 range} & \textbf{Count} \\
\hline
A & Infectious and Parasitic Diseases & 001-139 & 487 \\
B & Neoplasms & 140-239 & 892 \\
C & Endocrine, Nutritional, Metabolic & 240-279 & 743 \\
D & Blood and Blood-Forming Organs & 280-289 & 312 \\
E & Mental Disorders & 290-319 & 651 \\
F & Nervous System & 320-389 & 568 \\
G & Sense Organs & 390-459 & 89 \\
H & Circulatory System & 460-519 & 1000 \\
I & Respiratory System & 520-579 & 1000 \\
J & Digestive System & 580-629 & 897 \\
K & Genitourinary System & 630-679 & 734 \\
L & Pregnancy, Childbirth, Puerperium & 680-709 & 124 \\
M & Skin and Subcutaneous Tissue & 710-739 & 203 \\
N & Musculoskeletal System & 740-759 & 456 \\
O & Congenital Anomalies & 760-779 & 78 \\
P & Perinatal Conditions & 780-799 & 198 \\
Q & Symptoms and Ill-Defined Conditions & 800-999 & 785 \\
\hline
\textbf{Total} & & & \textbf{8,617} \\
\hline
\end{tabular}
\caption{ICD-9-CM disease chapters, code ranges, and per-chapter sample sizes in the full manifest. The majority-class baseline (Circulatory and Respiratory at 1{,}000 notes each) is 11.6\%.}
\label{tab:icd9-chapters}
\end{table}

\section{Models and dictionaries}
\label{app:models}

Table~\ref{tab:model-ids} gives the exact model identifiers and SAE specifications for the eleven models of \S\ref{sec:foundation}. The models span 34 to 64 layers and residual dimensions of 2{,}560 to 5{,}376. The MedGemma variants use the dictionaries trained on their Gemma-3 parents (the borrowed-parent configuration discussed in \S\ref{sec:existence}), and the fresh dictionary for II-Medical-8B was trained on that model's own activations at layer 24 with the parent architecture (TopK, 65{,}536 features, $K = 100$), reaching 97.4\% explained variance with 0.11\% dead features.

\begin{table}[h]
\centering
\small
\begin{tabular}{lllr}
\hline
\textbf{Model} & \textbf{HuggingFace ID} & \textbf{SAE source} & \textbf{Dictionary size} \\
\hline
Qwen3-8B-Base & Qwen/Qwen3-8B-Base & Qwen/SAE-Res-Qwen3-8B-Base-W64K-L0\_100 & 65,536 \\
Qwen3.5-9B-Base & Qwen/Qwen3.5-9B-Base & Qwen/SAE-Res-Qwen3.5-9B-Base-W64K-L0\_100 & 65,536 \\
II-Medical-8B & internlm/II-Medical-8B & Qwen/SAE-Res-Qwen3-8B-Base-W64K-L0\_100 & 65,536 \\
Qwen3.5-27B & Qwen/Qwen3.5-27B & Qwen/SAE-Res-Qwen3.5-27B-W80K-L0\_100 & 81,920 \\
Qwen3-30B-A3B-Base & Qwen/Qwen3-30B-A3B-Base & Qwen/SAE-Res-Qwen3-30B-A3B-Base-W128K-L0\_100 & 131,072 \\
Gemma-3-4B-pt & google/gemma-3-4b-pt & gemma-scope-2-4b-pt-res-all & 16,384 \\
Gemma-3-4B-it & google/gemma-3-4b-it & gemma-scope-2-4b-it-res-all & 16,384 \\
Gemma-3-27B-pt & google/gemma-3-27b-pt & gemma-scope-2-27b-pt-res-all & 16,384 \\
Gemma-3-27B-it & google/gemma-3-27b-it & gemma-scope-2-27b-it-res-all & 16,384 \\
MedGemma-4B-it & google/medgemma-4b-it & gemma-scope-2-4b-it-res-all & 16,384 \\
MedGemma-27B-text-it & google/medgemma-27b-text-it & gemma-scope-2-27b-pt-res-all & 16,384 \\
\hline
\end{tabular}
\caption{Model and dictionary identifiers. MedGemma dictionaries are borrowed from the corresponding Gemma-3 parents.}
\label{tab:model-ids}
\end{table}

\section{Existence: held-out separability and split controls}
\label{app:existence}

This appendix carries the split-control detail behind \S\ref{sec:existence}. Table~\ref{tab:held_out_sep} compares in-sample against held-out one-vs-rest AUC at the selected peak layer. Optimism (in-sample minus held-out) stays below 0.009 in every case and is negative on four of five Qwen models, meaning the held-out AUC often exceeds the in-sample value, a conservative outcome. The patient-level split, which forces all notes of a patient into the same half, guards against within-patient autocorrelation and leaves held-out AUC above 0.877, so the directions generalize across patients and not merely across notes of patients seen in selection.

\begin{table}[h]
\caption{In-sample against held-out one-vs-rest AUC at the selected peak layer. Optimism is in-sample minus held-out. The patient-level split groups all notes of a patient into the same half.}
\label{tab:held_out_sep}
\begin{center}
\begin{tabular}{llccc}
\toprule
\textbf{Model} & \textbf{Split} & \textbf{In-sample} & \textbf{Held-out} & \textbf{Optimism} \\
\midrule
Qwen3-8B-Base (L23) & Item & 0.8862 & 0.8814 & +0.005 \\
II-Medical-8B (L24) & Item & 0.8868 & 0.8810 & +0.006 \\
Qwen3-8B-Base (L23) & Patient (1,997 patients) & 0.8854 & 0.8772 & +0.008 \\
\bottomrule
\end{tabular}
\end{center}
\end{table}

The dictionary-independence check of \S\ref{sec:existence} uses the fresh II-Medical-8B dictionary of Appendix~\ref{app:models}. Projected into residual coordinates, its chapter directions cosine at $+0.835$ on average with those of the borrowed parent dictionary, positive on all seventeen chapters with range $+0.784$ to $+0.865$, so two dictionaries trained on different activation distributions agree on where the chapter lives.

\section{Clinical specificity: full firing-rate results}
\label{app:specificity}

Table~\ref{tab:firing_rates} reports the condition-wise firing rates behind \S\ref{sec:specificity} for the Qwen family. For TopK dictionaries a feature counts as active when it appears among the top-$K$ features of the token, and for JumpReLU dictionaries when its pre-activation exceeds the learned threshold. Full notes activate all ten selected features by construction, since the features are chosen as the ten highest selection-half AUC features on discharge notes. The Gemma family follows the same qualitative pattern.

\begin{table}[h]
\caption{Mean number of top-10 chapter features activating per condition (of 10), averaged over the seventeen chapters, Qwen family. Non-clinical text is fifteen short sentences on geography, sports, and food.}
\label{tab:firing_rates}
\begin{center}
\begin{tabular}{lccccc}
\toprule
\textbf{Model} & \textbf{Full notes} & \textbf{Bare Dx} & \textbf{Chapter title} & \textbf{Adjacent} & \textbf{Non-clinical} \\
\midrule
Qwen3-8B-Base     & 10.0 & 2.7 & --- & 0.6 & 0.013 \\
Qwen3.5-9B-Base   & 10.0 & 2.6 & 2.4 & 0.3 & 0.000 \\
Qwen3.5-27B       & 10.0 & 2.7 & 2.9 & 0.3 & 0.008 \\
Qwen3-30B-A3B-Base & 10.0 & 2.0 & 1.4 & 0.4 & 0.016 \\
II-Medical-8B     & 10.0 & 2.1 & 1.3 & 0.4 & 0.004 \\
\bottomrule
\end{tabular}
\end{center}
\end{table}

\section{Causal intervention: sweep detail and controls}
\label{app:causal_detail}

\paragraph{Magnitudes behind Table~\ref{tab:intervention}.} Across the four Qwen models with sound baselines, held-out $\Delta P$ at the selected configuration is $+0.041$ to $+0.069$, a relative lift of $+7\%$ to $+18\%$ over baselines of 0.376--0.581 where uniform chance is 5.9\%. II-Medical-8B's near-degenerate baseline can be attributed to post-training issues (a performance of 0.111, with the model answering one fixed letter for most notes) limits the interpretation of its row, and Appendix~\ref{app:roles} treats it separately. The Gemma effects reach $+0.0193$. On the specificity side, the matched-magnitude random perturbation is actively destructive on the pretrained Gemma models (mean $-0.035$ on the 4B and $-0.032$ on the 27B across seeds) while the chapter direction helps, so the models tolerate and use their own concept geometry at doses that break them along arbitrary directions. 

\begin{figure}[h]
\centering
\includegraphics[width=0.90\linewidth]{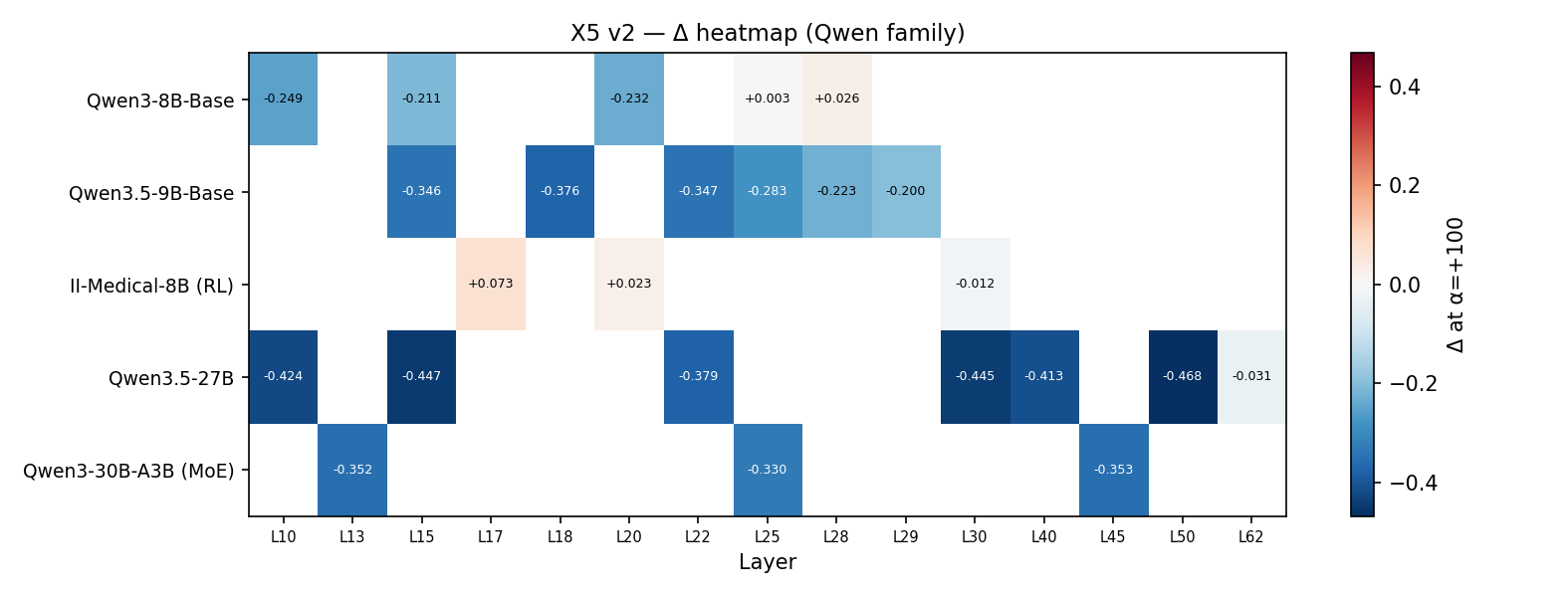}\\[2pt]
\includegraphics[width=0.90\linewidth]{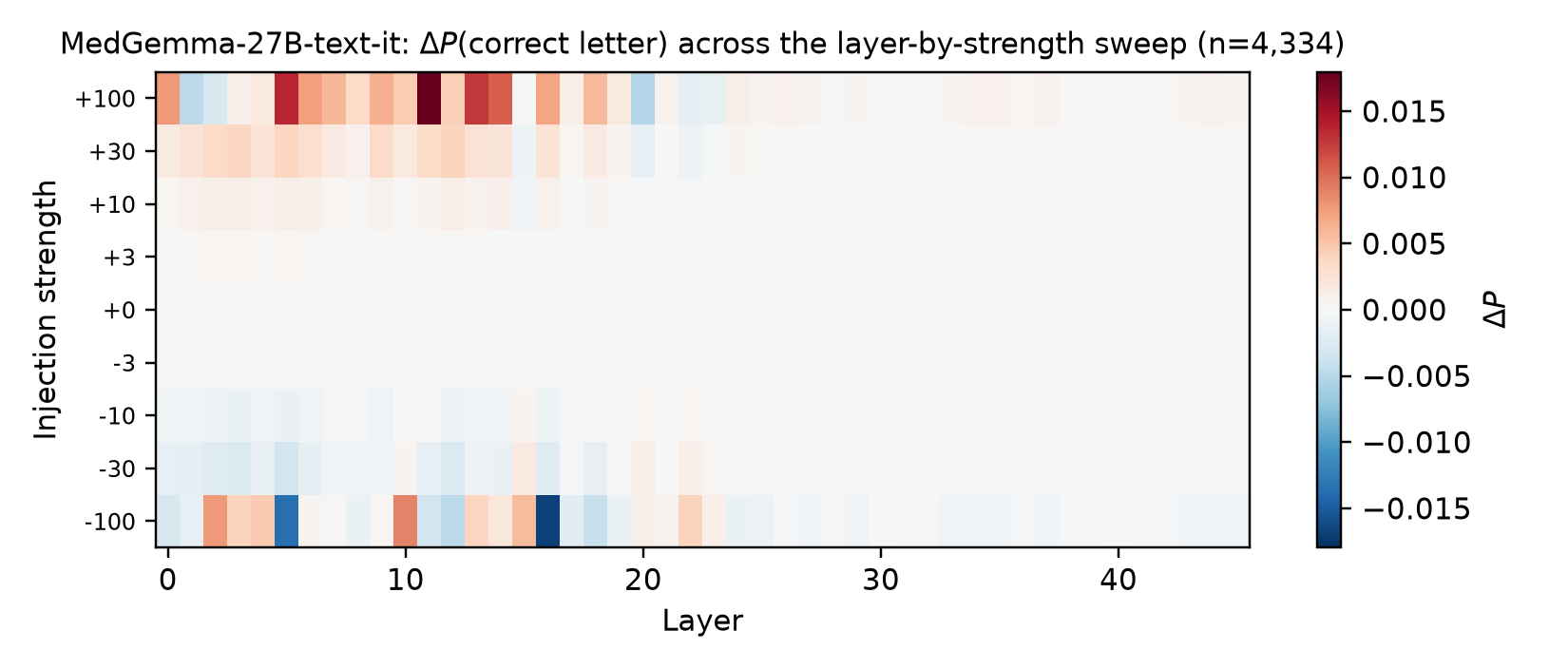}
\caption{$\Delta P(\text{correct letter})$ over the full layer-by-strength sweep, one model per family: Qwen3-8B-Base (top, held-out evaluation of the selection-half choice) and MedGemma-27B-text-it (bottom, full-manifest sweep). In both families the positive response concentrates in a band of layers even though separability is nearly uniform across depth, and grows with strength up to the swept maximum, the depth dissociation of \S\ref{sec:causal}.}
\label{fig:layer_heatmap}
\end{figure}

\paragraph{Sweep and controls.} Figure~\ref{fig:layer_heatmap} shows the full layer-by-strength sweep behind \S\ref{sec:causal} for Qwen3-8B-Base. The positive ridge concentrates in a band of layers even though the AUC profile of \S\ref{sec:existence} is nearly flat across depth, which is the depth dissociation of the main text. The layers around the marginal AUC maximum produce a behavioral swing of only 0.022 between the weakest and strongest injection, while earlier layers with essentially the same separability swing 0.283 under identical conditions, a fifteen-fold difference on the same model and direction. The unembedding control extracts the model's unembedding row for the correct answer letter, mean-centers and unit-normalizes it, and injects it at the same layer and strength, and its near-zero cosine with the chapter directions ($+0.003$ on Qwen3-8B-Base at layer 23) together with its failure to reproduce the effect rules out a direct lexical pathway. Benjamini-Hochberg correction at $q = 0.05$ is applied across the roughly 1{,}500 layer-strength combinations per model rather than across the selected configurations only.

The Gemma rows of Table~\ref{tab:intervention} follow the identical held-out protocol as the Qwen rows, a note-level split with the configuration chosen on the selection half over the positive strengths of the full layer-by-strength sweep and the effect evaluated on the held-out half. Selection optimism on the Gemma rows is between $-0.003$ and $+0.004$. The Gemma, Qwen3.5-27B, and mixture-of-experts chapter-specific values subtract the mean random-direction effect over seeds 42, 7, and 123 at the reported configuration (random effects per model: $-0.035$ on Gemma-3-4B-pt, $+0.005$ on Gemma-3-4B-it, $+0.009$ on MedGemma-4B-it, $-0.032$ on Gemma-3-27B-pt, $+0.001$ on Gemma-3-27B-it, $-0.006$ on MedGemma-27B-text-it, $-0.017$ on Qwen3.5-27B, and $-0.025$ on Qwen3-30B-A3B-Base, whose routing makes arbitrary perturbations costly while the chapter direction helps at the same dose). The remaining Qwen chapter-specific values were measured at each model's earlier best configuration, within a few layers of the reported one. For the MedGemma variants the dictionaries are trained on the Gemma-3 parents rather than on the fine-tuned models themselves, so the model's own medical training additionally shapes how the dictionary interacts with the activations, which is what the MedGemma-4B-it specificity value reflects, the instrument rather than the model. Training dictionaries natively on these variants is the straightforward extension and is left for future work.

\section{Register transfer}
\label{app:register}

Table~\ref{tab:transfer} carries the full transfer results behind \S\ref{sec:existence}. Directions fitted on discharge summaries are applied without refitting to note excerpts from a patient-question corpus with a different length distribution and voice. Retention normalizes by the in-corpus ceiling, $\mathrm{Retention} = (\mathrm{AUC}_{\text{out}} - 0.5)/(\mathrm{AUC}_{\text{ceiling}} - 0.5)$, because AUC has a floor of 0.5 rather than 0. Every Qwen entry reaches permutation $p < 1/2{,}000$ under a 2{,}000-shuffle null.

\begin{table}[h]
\caption{Out-of-register chapter separability, Qwen family. Directions are applied to the new register without refitting.}
\label{tab:transfer}
\begin{center}
\begin{tabular}{lcccc}
\toprule
\textbf{Model} & \textbf{Out-of-register AUC} & \textbf{In-corpus ceiling} & \textbf{Retention} & \textbf{Permutation $p$} \\
\midrule
Qwen3.5-27B          & 0.815 & 0.868 & 86\% & $<1/2{,}000$ \\
Qwen3.5-9B-Base      & 0.756 & 0.845 & 74\% & $<1/2{,}000$ \\
Qwen3-30B-A3B-Base   & 0.709 & 0.823 & 65\% & $<1/2{,}000$ \\
II-Medical-8B        & 0.646 & 0.691 & 76\% & $<1/2{,}000$ \\
Qwen3-8B-Base        & 0.630 & 0.792 & 45\% & $<1/2{,}000$ \\
\bottomrule
\end{tabular}
\end{center}
\end{table}

\section{Temporal persistence: measurement detail}
\label{app:temporal}

The 170 passes per model of \S\ref{sec:temporal} are ten held-out notes per chapter. Base models continue the discharge note itself. Chat models end their generations early under the bare continuation, so they are prompted for a detailed clinical assessment of the note covering the primary diagnosis, the differential, and the management plan, and both registers show the same sustained dominance shape. Generations run to a natural ending capped at 2{,}000 tokens, and most models write past 1{,}200 tokens at the median. II-Medical-8B stops at a median of 359 tokens, which is why its curve ends near token 500.

Each model is read at its most responsive layer among those with near-complete finite coverage (at least 95\% of the best layer's finite-token count). The coverage condition matters on the two 27B Gemma dictionaries, whose last layers produce non-finite codes concentrated at early token positions, so an unconstrained effect-size maximum would select a layer that cannot even be evaluated over the opening tokens. The underlying projection grows monotonically with depth (a 130-fold increase from the first to the last layer on Qwen3-8B-Base), consistent with the depth picture of \S\ref{sec:causal}. II-Medical-8B is the instructive per-model case, near zero over its opening tokens and rising later in the window, consistent with its spikier, content-locked signal, where the projection lights up on chapter-aligned diagnostic tokens.

\section{Role priming: six-condition design and the degraded-model rescue}
\label{app:roles}

\begin{table}[h]
\caption{An assigned specialty moves the model through its concept centers, on both families (deferred from \S\ref{sec:roles_results} for space). Paired $t$ compares matched against counterfactual specialty on true-chapter projections, and accuracy gain is matched specialty over the no-role baseline (paired sign test).}
\label{tab:rolepriming}
\begin{center}
\small
\begin{tabular}{lccc}
\toprule
\textbf{Model} & \textbf{Paired $t$ (matched vs.\ counterfactual)} & \textbf{Acc.\ gain (pp)} & \textbf{Sign-test $p$} \\
\midrule
Qwen3-8B-Base      & +9.18 & +5.88 & $1.8 \times 10^{-4}$ \\
Qwen3.5-9B-Base    & +8.92 & +4.12 & $2.6 \times 10^{-3}$ \\
Qwen3.5-27B        & +8.63 & +3.82 & $2.4 \times 10^{-4}$ \\
Qwen3-30B-A3B-Base & +9.53 & +2.94 & $0.087$ \\
\midrule
Gemma-3-4B-pt        & +3.70 & +7.65  & $8.7 \times 10^{-7}$ \\
Gemma-3-4B-it        & +2.90 & +8.24  & $5.8 \times 10^{-8}$ \\
MedGemma-4B-it       & +5.59 & +7.35  & $4.2 \times 10^{-7}$ \\
Gemma-3-27B-pt       & +3.12 & +11.18 & $4.1 \times 10^{-10}$ \\
Gemma-3-27B-it       & +5.92 & +4.71  & $1.4 \times 10^{-4}$ \\
MedGemma-27B-text-it & +6.80 & +5.88  & $1.9 \times 10^{-6}$ \\
\bottomrule
\end{tabular}
\end{center}
\end{table}

The six conditions of \S\ref{sec:roles} are matched specialty, counterfactual specialty, randomly drawn clinical specialty, generalist physician, non-clinical role, and no role, giving 105 role prompts per model over the seventeen chapters plus global controls. The decisive contrasts on Qwen3-8B-Base are random-clinical minus baseline at $-0.0815$ (paired $t = -0.22$, $p = 0.83$, a randomly assigned clinical specialty does nothing), matched minus random-clinical at $+1.3010$ ($t = +9.57$, $p = 2.3\times10^{-19}$, 14 of 17 chapters), and matched minus generalist at $+0.6939$ ($p = 0.054$), placing the unspecified-physician condition between the two, as expected if the effect tracks specialty identity rather than clinical register. The matched-over-random contrast exceeds $t = 7.6$ on four of five models, the accuracy gain of Table~\ref{tab:rolepriming} holds on 15--16 of 17 chapters per Qwen model, and the counterfactual condition stays within 0.3 accuracy points of the no-role baseline on the Qwen models and within about two points on the Gemma instruct models. Projections in Table~\ref{tab:rolepriming} are read at each model's best persona-effect layer. Reading depth matters, since the persona-effect peak moves from 83\% of depth on Qwen3-8B-Base through 91\% (9B) and 94\% (30B-A3B) to 97\% (27B), while the chapter-encoding peak moves the opposite way with scale (roughly 15\% of depth on the 27B model against 85\% on the 8B), so any fixed-depth evaluation will miss one of the two signals on most models. The chat-template comparison holds the model fixed and toggles only the template, and the persona-specific mean shift is $3.5\times$ larger without the template on Qwen3.5-27B.

\paragraph{Measurement caveat for role-priming analyses.} For those interested in performing this kind of latent-space evaluation, it is important to note that the chat template dilutes the persona signal at the default final-position read. A paired within-model comparison on Qwen3.5-27B, identical prompts with and without the template, shows a $3.5\times$ larger persona-specific mean shift without it, so instruct-tuned deployments carry the same internal effect while showing less of it at the position most evaluations read, and analyses that read only there will underestimate role influence. This is important since the final tokens for the instruct models relate more to formatting and instruction/chat templating instead of the actual clinical content.

\paragraph{Rescuing a degraded model through its concept centers.} The medical fine-tune II-Medical-8B is the case where the language substrate and the latent space disagree outright, and it doubles as direct evidence for the depth picture of \S\ref{sec:causal}. Under the constrained-letter task the model answers one fixed letter for roughly 98\% of notes and scores 6.2\%, barely above chance, yet its internal chapter separation at the pre-answer position is the strongest we measure anywhere (Welch $t = +40.5$, mean projection $+74$ on the true chapter against $-4.9$ on wrong ones). The model knows and cannot say, an extreme clinical instance of the internal-knowledge gap \citet{orgad2025llms} documented in general domains. From the language substrate this model looks broken, but mechanistically it is a model with intact clinical concepts and a damaged output mapping, and the causal-depth picture says exactly where to intervene, at the layers that govern how the latent computation becomes the response. Injecting the chapter direction there more than doubles the probability of the correct answer, from 0.062 at baseline to 0.135 at the optimum, with the chapter-specific effect $+0.057$ against a five-seed random floor of $+0.016$, roughly thirty-five times the random recovery, converging with the triage-failure localization of \citet{navarro2026internal} from the causal side. The useful operating range is wide, and past the optimum the rescue degrades gradually while never falling below baseline within the swept range. The raw difference-of-means direction also helps ($+0.033$ chapter-specific) but is about 70\% weaker, consistent with the axis comparison of Appendix~\ref{app:sae_vs_raw}.

\section{SAE-decoder against raw difference-of-means as the steering axis}
\label{app:sae_vs_raw}

Table~\ref{tab:sae_vs_raw} gives the dose curves behind the axis choice of \S\ref{sec:steering}, computed on identical models, layers, notes, and doses, so the only moving part is the direction. On Qwen3-8B-Base the raw direction produces no positive dose anywhere on its curve while the SAE-decoder direction peaks at $+0.038$, on II-Medical-8B both rescue the degraded output but the SAE-decoder direction is about 70\% more effective, and on the mixture-of-experts Qwen3-30B-A3B the raw direction is destructive at every positive dose tested while the SAE-decoder direction still helps at small doses, with the mixture-of-experts model overall needing far smaller doses than the dense models, consistent with its routing being easier to knock off distribution. Our reading in the main text follows from these curves, the decoder reconstruction concentrates the direction on feature axes that downstream computation consumes, while the raw mean also carries components that correlate with the chapter without taking part in the computation.

\begin{table}[h]
\caption{Change in correct-letter probability against dose for the two steering axes at each model's selected layer. Same notes, layers, and doses within each block.}
\label{tab:sae_vs_raw}
\begin{center}
\small
\begin{tabular}{lcc@{\hskip 2em}cc@{\hskip 2em}cc}
\toprule
 & \multicolumn{2}{c}{\textbf{Qwen3-8B-Base (L25)}} & \multicolumn{2}{c}{\textbf{II-Medical-8B (L17)}} & \multicolumn{2}{c}{\textbf{Qwen3-30B-A3B (L45)}} \\
$\alpha$ & raw & SAE-dec & raw & SAE-dec & raw & SAE-dec \\
\midrule
+3   & +0.002 & +0.006 & +0.001 & +0.001 & ---    & ---    \\
+10  & +0.003 & +0.017 & ---    & ---    & $-$0.029 & +0.009 \\
+30  & $-$0.005 & \textbf{+0.038} & +0.025 & +0.030 & $-$0.153 & $-$0.053 \\
+100 & $-$0.159 & +0.003 & +0.049 & \textbf{+0.073} & $-$0.357 & $-$0.353 \\
\bottomrule
\end{tabular}
\end{center}
\end{table}

\section{Dose calibration for generation steering}
\label{app:dose}

The relative-dose rule of \S\ref{sec:steering} matters because residual-stream norms differ by orders of magnitude across models and grow with depth. As a calibration illustration, on Gemma-3-4B-it, whose residual scale is the largest in our pool (median chapter-contrast norm 666.6 at the steered layer against 18.1 for Qwen3-8B-Base), an absolute dose that steers the 8B Qwen model amounts to $0.02\times$ of the natural between-chapter contrast and changes 13 of 1{,}624 generated answers against 12 for the random control, an intervention that never fired, while the matched relative dose $\rho = 0.55$ yields $+0.0123$ with 118 discordant answers and the predicted result ordering (Table~\ref{tab:generation_steering}). The rule for any cross-model steering comparison is therefore twofold, express the dose as a fraction of the model's own median chapter-contrast norm at the steered layer, and report the discordance count next to every effect so that an intervention that never fired cannot be read as a null.

\begin{table}[h]
\caption{Steering under the deployment prior (deferred from \S\ref{sec:gensteer} for space; injection held active through generation, 1{,}624 held-out notes per model). $\Delta$~Frac.\ is the change in the fraction of generated diagnoses naming the correct chapter under the true-chapter direction, and Discordant counts how many answers changed at all. Doses are relative ($\rho$ of the model's median chapter-contrast norm at the steered layer).}
\label{tab:generation_steering}
\begin{center}
\begin{tabular}{lccccc}
\toprule
\textbf{Model} & \textbf{$\rho$} & \textbf{$\Delta$ Frac.} & \textbf{$p$} & \textbf{Discordant} & \textbf{Ordering} \\
\midrule
Qwen3-8B-Base   & 0.55--1.77 & +0.0111 & 0.047 & 74  & true $>$ rand $>$ wrong \\
Qwen3.5-9B-Base & 0.55--1.77 & +0.0234 & 0.001 & 136 & true $>$ rand $>$ wrong \\
Gemma-3-4B-it   & 0.55       & +0.0123 & 0.080 & 118 & true $>$ rand $>$ wrong \\
\bottomrule
\end{tabular}
\end{center}
\end{table}

Two methodological choices thus define the steering appraoch for \S\ref{sec:gensteer}. First, the dose is relative to the model's own representation scale, a fraction $\rho$ of the model's typical between-chapter delta at the steered layer, since residual-stream norms vary by orders of magnitude across models and depths. We report discordance counts (how many outputs changed at all) next to every effect so that an intervention that never fired cannot be mistaken for a null result. Second, the main steering axis is the SAE-decoder direction of Equation~\ref{eq:sae_direction} and not the raw difference of means. In our ablations, on identical models, layers, and doses, the SAE direction wins every head-to-head comparison we run. 
Intuitively, this can be explained by the decoder reconstruction expressing the concept along the feature axes the downstream layers actually read, while the raw difference of means also finds components that merely correlate with the chapter without taking part in the latent computation, so part of its dose is spent on directions which the model never uses.

\section{Clinician study protocol}
\label{app:clinician_protocol}

\paragraph{Cohort.} HealthBench conversations \citep{arora2025healthbench} are drawn from the consensus, hard, and professional strata, sampled in that priority order, and filtered to a chapter-faithful orthopedic cohort matched to the Musculoskeletal chapter. Chapter faithfulness excludes look-alike neighbors (fractures belong to the Injury chapter and gout to Endocrine, so conversations about them are excluded), red-teaming items are excluded as a different construct, and non-English conversations are removed with a language screen whose short-text false positives are corrected by an English stopword-density rule. The cohort holds about one hundred conversations.

\paragraph{Conditions and generation.} Each conversation is answered by Qwen3.5-27B under three conditions, no role, matched-specialty system role, and adversarial other-specialty system role. Generation uses the benchmark's own sampling parameters, all conditions of an item share one random seed so that divergence between responses is attributable to the condition and not to sampling, responses run to a natural ending (items that reach the token cap are regenerated at a doubled cap under the same seed, which reproduces the identical prefix and completes it), and reasoning-trace modes are disabled so raters see final responses only. Every generation records the per-token SAE feature firings at the representation-peak and steering layers and the dense projections onto all seventeen chapter directions at every layer. 

\paragraph{Screening and blinding.} Cases enter workbooks with response positions randomized per row, condition identities absent, three pairwise judgments per row with an explicit tie option, and a fingerprint that binds each workbook to its sealed answer key. The final workbook carries 41 Qwen3.5-27B cases.

\paragraph{Frozen predictions.} Before any rating, every response's per-token trace at the representation-peak layer is reduced to the mean target-chapter dominance of \S\ref{sec:clinician}, and the frozen rule predicts the higher-dominance response of every pair, with the signed activation gap of every pair stored. Planned analysis on return of the ratings is agreement of blinded preference with the frozen predictions via sign tests on non-tie judgments, stratified by comparison type and model, and inter-rater agreement on the double-rated subset. 
% Secondary scalars (occupancy, peak dominance, and intrusion of the opposing chapter) are stored for exploratory analysis and make no prediction.

\paragraph{Raters, returns, and integrity.} The workbooks were completed independently by a senior orthopaedic resident surgeon and a junior medical student, giving 246 head-to-head judgments over the 41 Qwen3.5-27B cases plus a free-text comment field per case. Before analysis, every displayed response in the returned workbooks was re-verified against the stored generation of its keyed (condition, case) by content fingerprint, all matching, so no judgment can be attributed to the wrong condition.

\paragraph{Capability exclusions.} Three cases were excluded from every computed statistic because the failures the raters reported trace to a capability the model was never given. The pattern is consistent, since these conversations ask the model about specific literature (one asks about a specific named review, one asks for comparative evidence across three injection-guidance techniques, one asks for the best surgical approach and drew responses citing ``highly specific recent studies that ... appear fabricated''), and a model without web access can only fabricate such references. The senior's comment on the evidence-comparison case states the problem plainly, ``Fabricated references in all. Although, B got 2 of the references correct,'' which means ratings on these cases measure verification impossibility rather than response quality. The exclusion list with per-case reasons and curated copies of both workbooks with the exclusions applied are released with the data, and all analysis code reproduces every number in this appendix from those files.

\paragraph{Results in full.} After exclusions there are 228 paired judgments over 38 cases, with 34 ties in the pooled set, and ties are informative rather than noise, since raters used them exactly where responses converge. The per-rater agreement of the activation-strength prediction is 57.9\% for the junior ($n = 95$) and 53.5\% for the senior ($n = 99$), and the gap-stratified values in \S\ref{sec:clinician} pool both raters. Inter-rater agreement over the three-way choice including Tie is raw 0.58 with Cohen's kappa 0.43, moderate agreement between two clinicians at very different career stages who never saw each other's sheets.

\paragraph{Agreement by activation gap.} Table~\ref{tab:gap_agreement} carries the counts and tests behind the main-text percentages, over all pairs and stratified by comparison type. The stratification shows the prediction is not a relabeling of the priming condition. Among plain-versus-adversarial pairs alone it reaches 64.3\% overall and 77.8\% on the larger-gap half, while aligned-versus-plain pairs give it almost nothing to separate (median activation gap 0.048 against 0.082 for plain-versus-adversarial), and there it stays at chance, with the small gap known before any rating comes back.

\begin{table}[h]
\centering
\small
\caption{Agreement of the activation-strength prediction with blinded clinician preference over non-tie judgments, overall and within the half of pairs whose activation gaps are larger (threshold at each subset's median gap, ties at the threshold included). $p$ is the exact two-sided binomial test against 0.5. On the quarter of pairs with the largest gaps the pooled agreement reaches 76.0\% (38 of 50, $p = 3.1\times10^{-4}$). Gaps are $|\mathrm{dom}_A - \mathrm{dom}_B|$ from the frozen predictions.}
\label{tab:gap_agreement}
\begin{tabular}{lccccccc}
\toprule
& & \multicolumn{3}{c}{\textbf{All pairs}} & \multicolumn{3}{c}{\textbf{Larger-gap half}} \\
\cmidrule(lr){3-5}\cmidrule(lr){6-8}
\textbf{Pairs} & \textbf{Median gap} & Agree & $n$ & $p$ & Agree & $n$ & $p$ \\
\midrule
All comparisons          & 0.058 & 55.7\% & 194 & 0.13  & 66.0\% & 97 & 0.002 \\
Plain vs.\ adversarial   & 0.082 & 64.3\% & 70  & 0.022 & 77.8\% & 36 & 0.001 \\
Adversarial vs.\ aligned & 0.057 & 53.3\% & 60  & 0.70  & 66.7\% & 30 & 0.099 \\
Aligned vs.\ plain       & 0.048 & 48.4\% & 64  & 0.90  & 54.5\% & 33 & 0.73  \\
\bottomrule
\end{tabular}
\end{table}

\paragraph{Condition-level comparisons (secondary finding).} The three priming conditions exist to induce different behaviors on the same case so that the mechanistic signal has variation to predict, and the condition-level preferences are therefore a by-product of the design rather than its aim. Table~\ref{tab:clinician} reports them in full. The plain model beats the adversarially primed one 50 to 20 among non-tie judgments (exact binomial $p = 4.4\times10^{-4}$), each rater landing on exactly 25 to 10 independently ($p = 0.017$) with the consensus subset at 20 to 6 ($p = 0.009$), and the plain model also beats the matched-specialty priming, 41 to 23 ($p = 0.033$), the open-ended counterpart of the closed-task gains of \S\ref{sec:roles}. The comments say where the two settings diverge, because the persona buys diagnostic focus at the price of invented authority, as documented in the qualitative themes below. The relevant point for the paper is that neither direction of priming tracks quality the way its prompt semantics promise, while the concept-center activations track it wherever they separate the responses, which is the main-text finding of \S\ref{sec:clinician}.

\begin{table}[h]
\caption{Blinded clinician preferences on the orthopedic cohort (Qwen3.5-27B), a secondary finding. Wins--losses among non-tie judgments with exact binomial $p$ values. Consensus counts only comparisons where both raters chose the same option.}
\label{tab:clinician}
\begin{center}
\small
\begin{tabular}{lcccc}
\toprule
\textbf{Comparison} & \textbf{Junior} & \textbf{Senior} & \textbf{Consensus} & \textbf{Aggregate} \\
\midrule
unsteered $>$ adversarial & 25--10 (.017) & 25--10 (.017) & 20--6 (.009) & 50--20 (.0004) \\
unsteered $>$ aligned     & 22--10 (.050) & 19--13 (.377) & 13--5 (.096) & 41--23 (.033) \\
aligned $>$ adversarial   & 19--9 (.087)  & 17--15 (.860) & 10--5 (.302) & 36--24 (.155) \\
\bottomrule
\end{tabular}
\end{center}
\end{table}

\paragraph{Where the raters disagree, and why.} The raters disagree on 48 of the 114 paired comparisons, split between 20 tie-versus-choice mismatches and 28 opposite choices. Reading the comments on the disagreement rows shows the two raters weighting different things rather than reading the responses differently. The senior flags concrete technical faults and missing patient context that decide the comparison outright. On a case asking for a step-by-step technique to elevate a collapsed lateral tibial plateau, the junior chose the aligned response as ``the most useful surgical-technique response while retaining appropriate caveats,'' while the senior chose the plain one because ``there are technical flaws in the surgical procedure mentioned.'' On a case where a man describes his wife's intermittent facial grimacing, the junior preferred the aligned response as the one that ``most clearly covers the neurological red flags and escalation criteria,'' while the senior preferred the plain response because it ``is the only one that explicitly checks the wife's awareness.'' On an under-specified back-pain timeline the senior chose the plain response with the note ``As the case conversation is incomplete. Response C asking for more details is the best.'' The junior, in turn, weights completeness and guideline coverage (``A most closely follows the Jones-criteria/differential/uncertainty requirements''). The disagreements are therefore complementary readings of quality, the junior scoring coverage and the senior scoring correctness and context-seeking, which is also why the consensus subset, where both weightings agree, gives the clearest preference counts in Table~\ref{tab:clinician}.

\paragraph{Qualitative themes with the underlying text.} Three themes carry most of the signal, and each connects to the mechanistic story. First, persona-induced fabrication of specifics. Asked to draft an inpatient progress note from a trauma-ward recording, the aligned model invents a complete objective section (``**Vital Signs:** Blood Pressure: 128/78 mmHg, Heart Rate: 72 bpm ... Temperature: 98.4°F'') where the conversation contains no vitals at all, and the junior chose the plain response because it ``avoids inventing exact observations,'' while the adversarially primed model, answering an emergency physician's question about in-hospital cardiac arrests, confabulates an entire biography (``As a hematologist who has spent decades in the trenches of critical care and resuscitation science...''), marked as ``less reliable framing and persona fabrication.'' The persona instruction is a single sentence, and the model expands it into invented vitals or an invented career, the role pushing content the model cannot ground. Second, the adversarial persona drags the content toward its own specialty, which is the concept intrusion our traces measure, seen by the raters unprompted. On a six-month low-back-pain case with radiation to the leg, the adversarially primed response ``overemphasises malignancy/hematologic causes,'' and on a knee-replacement eligibility question the rater ranked the adversarial response last for its ``unnecessary hematology framing.'' Third, the roles cost calibration. On an adult-onset Still's-like presentation (daily fever spikes with a salmon-pink truncal rash), the plain response ``is the most appropriately cautious; B/C move toward treatment/confirmation based on insufficient information,'' and the same pattern repeats wherever the case under-determines the diagnosis. These comments explain the direction of Table~\ref{tab:clinician}, since both roles, aligned and adversarial alike, inject authority the model does not have, the adversarial one additionally injects its own specialty's content, and they also explain why the closed-task gains of \S\ref{sec:roles} do not carry over, because in a seventeen-way classification a persona can only move the answer, while in an open-ended response it also moves the framing, the claimed evidence, and the invented particulars that a clinician reads as untrustworthy.

\section{Extended discussion}
\label{app:discussion}

Across our experiments, one key insight is that the presence of a clinical concept inside a model is spread across its depth. There is a circuit across the layers that encodes the concept center, and it is used at every generation step. The concept is separable almost equally at every depth, yet only some depths are causally potent. The persona effect lives later than the chapter encoding, and it moves later still with scale, from 83\% of depth on the 8B model to 97\% on the 27B. The final token under a chat template hides what the mid-sequence positions show. The constrained answer responds to a targeted impulse at the injection point, while open-ended generation responds to a sustained one. This fits the staged-inference view of transformer depth \citep{lad2024remarkable} and the observation that intermediate layers carry the best task representations \citep{skean2025layer}, and it explains, without any appeal to model deficiency, why single-layer evaluations in this domain produce false nulls. The shape of our depth results also matches the staged description \citet{liu2025relevance} give of relevance judgment, where input information is extracted early, processed against the instruction in the middle layers, and converted into the required output by late attention heads. Along these lines, the broad early-to-middle band where the chapter concept stays separable is where the clinical semantics of the note are integrated into the model's reasoning, while the causally effective steering band, sitting at or beyond the middle of the network and reaching the final quarter in the largest models, is where the model decides how those latent computations reach the user. These findings pave the way for a new regime of clinical AI evaluations, with a focus on the latents of the models, which are increasingly becoming more sophisticated and internally coherent, during clinical decision support, and warrant a mechanistic mapping of their latent reasoning in addition to the language substrate. 

% The corpus is one hospital system's discharge summaries, and although the directions transfer to a patient-question register with most of their separability retained, broader registers are untested. Chapter granularity is deliberately coarse. The ongoing work the narrative points to extends the same representations rather than replacing them, injecting wrong reasoning early in generation and measuring recovery, and department-prior up- and down-weighting with the cost on non-matching tasks measured explicitly.

\end{document}